\documentclass[10pt]{SelfArx} 

\usepackage[english]{babel} 

\usepackage{moreverb}
\usepackage{footnote}
\usepackage{xcolor}
\usepackage{url}
\usepackage{bm} 
\usepackage{siunitx}
\usepackage[colorlinks,bookmarksopen,bookmarksnumbered,citecolor=red,urlcolor=red]{hyperref}
\usepackage[caption=false,font=footnotesize]{subfig}

\newcommand\BibTeX{{\rmfamily B\kern-.05em \textsc{i\kern-.025em b}\kern-.08em
T\kern-.1667em\lower.7ex\hbox{E}\kern-.125emX}}

\definecolor{color1}{RGB}{0,0,90} 
\definecolor{color2}{RGB}{0,20,20} 

\usepackage{hyperref} 
\hypersetup{hidelinks,colorlinks,breaklinks=true,urlcolor=color2,citecolor=color1,linkcolor=color1,bookmarksopen=false,pdftitle={Title},pdfauthor={Author}}

\JournalInfo{~} 
\Archive{} 

\PaperTitle{DREAM Architecture: a Developmental Approach to Open-Ended Learning in Robotics} 

\Authors{Stephane Doncieux\textsuperscript{1}*, Nicolas Bredeche\textsuperscript{1}, Leni K. Le Goff\textsuperscript{1}, Beno\^{i}t Girard\textsuperscript{1}, Alexandre Coninx\textsuperscript{1}, Olivier Sigaud\textsuperscript{1}, Mehdi Khamassi\textsuperscript{1}, Natalia Díaz-Rodríguez\textsuperscript{2}, David Filliat\textsuperscript{2}, Timothy Hospedales\textsuperscript{3}, A.E. Eiben\textsuperscript{4}, Richard Duro\textsuperscript{5}
} 


\newcommand{\keywordname}{Keywords} 

\affiliation{\textsuperscript{1} \textit{Sorbonne Universit\'e, CNRS, ISIR, Paris, France}}
\affiliation{\textsuperscript{2} \textit{U2IS, INRIA Flowers, ENSTA, Institut Polytechnique Paris}}
\affiliation{\textsuperscript{3} \textit{University of Edinburgh}}
\affiliation{\textsuperscript{4} \textit{Vrije Universiteit Amsterdam}}
\affiliation{\textsuperscript{5} \textit{GII, CITIC, Universidade da Coru\~na}}
\affiliation{*\textbf{Corresponding author}: stephane.doncieux@sorbonne-universite.fr} 

\begin{document}







\Abstract{Robots are still limited to controlled conditions, that the robot designer knows with enough details to endow the robot with the appropriate models or behaviors. Learning algorithms add some flexibility with the ability to discover the appropriate behavior given either some demonstrations or a reward to guide its exploration with a reinforcement learning algorithm. Reinforcement learning algorithms rely on the definition of state and action spaces that define reachable behaviors. Their adaptation capability critically depends on the representations of these spaces: small and discrete spaces result in fast learning while large and continuous spaces are challenging and either require a long training period or prevent the robot from converging to an appropriate behavior. Beside the operational cycle of policy execution and the learning cycle, which works at a slower time scale to acquire new policies, we introduce the \textit{redescription cycle}, a third cycle working at an even slower time scale to generate or adapt the required representations to the robot, its environment and the task. We introduce the challenges raised by this cycle and we present DREAM (Deferred Restructuring of Experience in Autonomous Machines), a developmental cognitive architecture to bootstrap this redescription process stage by stage, build new state representations with appropriate motivations, and transfer the acquired knowledge across domains or tasks or even across robots. We describe results obtained so far with this approach and end up with a discussion of the questions it raises in Neuroscience.
}


\maketitle

\section{Introduction}

\textit{What do we miss to build a versatile robot, able to solve tasks that are not pre-programmed, but that could be given on-the-fly and in an unprepared environment?} Robots with these capabilities would pave the way to many applications, from service robotics to space exploration. It would also reduce the need for deep analyses of a robot's future environments while designing it. 

Floor cleaning robots are the only autonomous robots that have been able to solve real-world problems out of the lab and have proved their efficiency on the market where they have been sold by millions. The variability of our everyday environments, that are unknown to the robot designers, is handled by a carefully tuned behavior-based architecture dedicated to this single, well-defined task \cite{jones2006robots}. But even in this case, and after years of improvement involving tests and upgrades by skillful engineers, a significant number of users finally stop using them because their behavior is not adapted to their home \cite{VAUSSARD2014376}. Dealing with variable environments is thus a challenge, even in this context, and users call for more adaptivity \cite{VAUSSARD2014376}. 

The adaptivity we will focus on here is the ability to solve a task when the appropriate behavior is not known beforehand. 
The goal-driven exploration strategy to learn a policy will be modeled as a reinforcement learning process. Reinforcement learning relies on a Markov Decision Process (MDP) that includes a state space, an action space, a transition function and a reward function \cite{sutton1998introduction}. Although widely used in machine learning, reinforcement learning is notoriously hard to apply in robotics as the definition of the relevant MDP critically impacts the learning performance and needs to be adapted to the task \cite{kober2013reinforcement}. This raises an issue: if the task is not known by the robot designer, it will not be possible to define an appropriate MDP at robot design time. A possibility is to rely on end-to-end learning \cite{levine2016end}, but even in this case, some careful preparation is required. Besides, it would be interesting to avoid starting from scratch for each new task and to transfer the previously acquired knowledge to a new context \cite{taylor2009transfer,pan2010survey,lazaric2012transfer,lu2015transfer,hospedales2020metalearning}. 

Reusing past experience is particularly important in robotics where the sampling cost is high: testing a policy may damage the robot if it does not respect safety constraints \cite{garcia2015comprehensive} and even if it is safe, it will increase the wear and tear of the robot.  Many different approaches have been proposed to reuse the already acquired experience \cite{fernandez2010probabilistic, munoz2016transfer, nguyen2017scalable, kaboli2018active}. Transfer learning is also particularly interesting to learn in simulation before transferring to reality, as it drastically reduces the number of required samples on the real robot \cite{cully2015robots,clavera2017policy, devin2017learning, james2017transferring,traore2019discorl}. Specific knowledge representations can clearly facilitate the transfer of acquired knowledge \cite{barreto2017successor}, thus the choice of an appropriate representation is also important for this question.

Human beings do not use a single representation to solve the problems they are facing. Their ability to build new representations even seems to be a critical factor of their versatility \cite{karmiloff1992beyond,redish2007,collins2012,tervo2016}. The features of an MDP representation constrains the kind of learning algorithms that can be used: a small and discrete set of actions and states facilitates an exhaustive exploration to discover the most relevant policy \cite{sutton1998introduction}, while a large dimension and continuous state and action spaces raise exploration issues that can be solved, for instance, by restraining exploration to the neighborhood of an expert demonstration, if available \cite{abbeel2010autonomous,mulling2013learning}, by endowing the agent with intrinsic motivation mechanisms \cite{Baranes2013goalexploration,colas2019curious}  or by combining fast and slow learning \cite{khamassi2012,caluwaerts12,BOTVINICK2019408,hospedales2020metalearning}. 

\textit{What if a robot could switch between different representations and build new ones on-the-fly?} An adapted representation would allow the robot to (1) understand a task, by  identifying the target and associating it with a state space that it can control or learn to control and (2) search for a solution. When the robot knows little about the environment and the task, it could use end-to-end strategies and switch to faster decision or learning processes based on adapted representations. Following the framework introduced in \cite{doncieux2018open}, it is assumed here that a single state and action space cannot cover all the tasks the robot may be confronted with. Therefore we go beyond a single task resolution and consider the acquisition of an appropriate representation as a challenge to be explicitly addressed. Besides, we consider the acquisition of new knowledge representations as a challenge per se, that may require specific processes that are not necessarily task-oriented. As human infants, the proposed approach needs to face the challenges of understanding the robot's environment and its own capabilities \cite{lungarella2003developmental,weng2004developmental,stoytchev2009some,cangelosi2015developmental}.

\begin{figure}[htb!]
\begin{center}
\includegraphics[width=\linewidth]{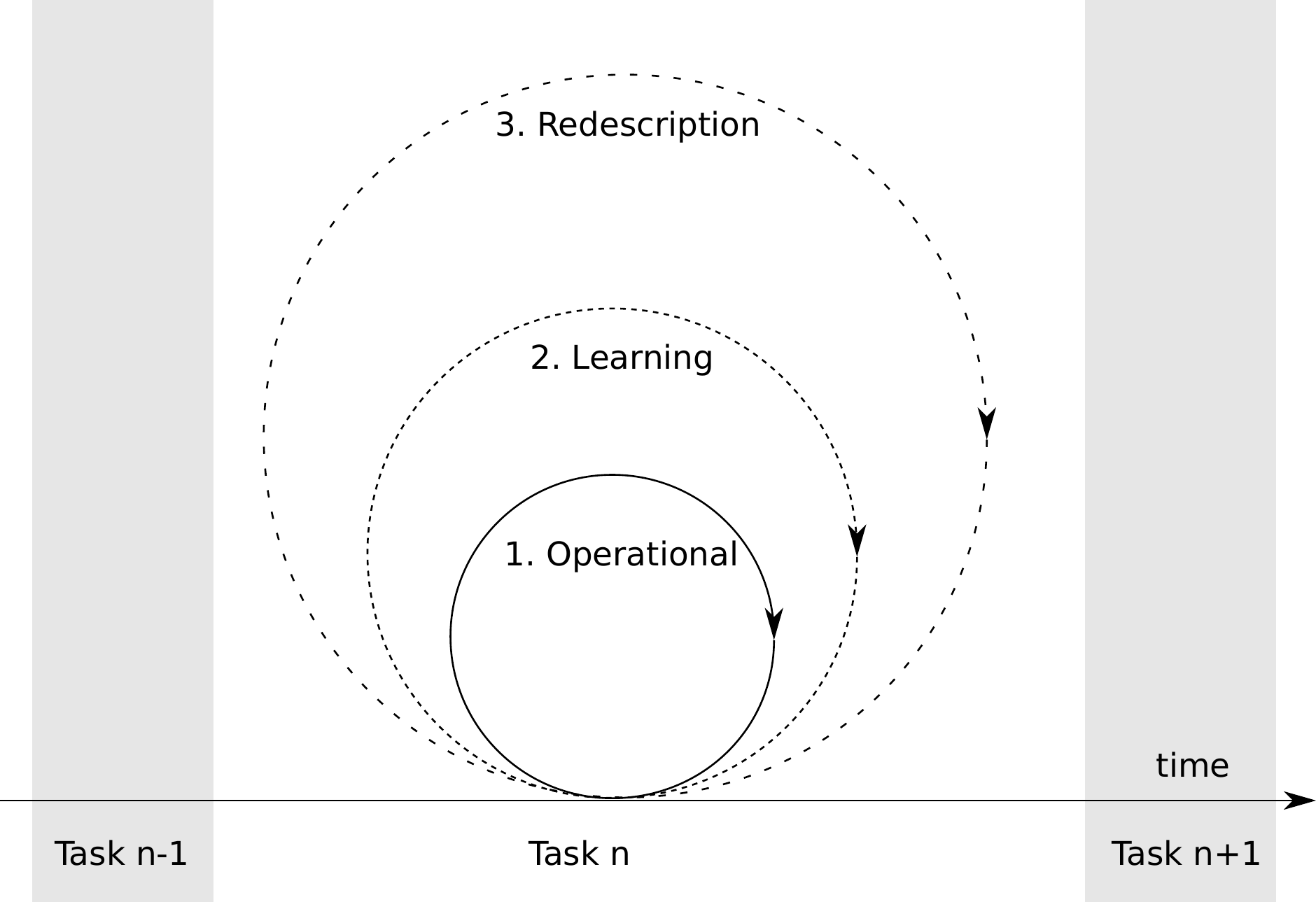}
\end{center}
\caption{\label{fig:3loops}The three required loops for a robot to be endowed with the ability to adapt to new environments: 1. Operational: the robot applies a known policy, i.e. a mapping from states to actions, 2. Learning: the robot acquires a new policy, 3. Redescription: the robot discovers new states and action spaces. }
\end{figure}

\begin{figure}[htb!]
\begin{center}
\includegraphics[width=\linewidth]{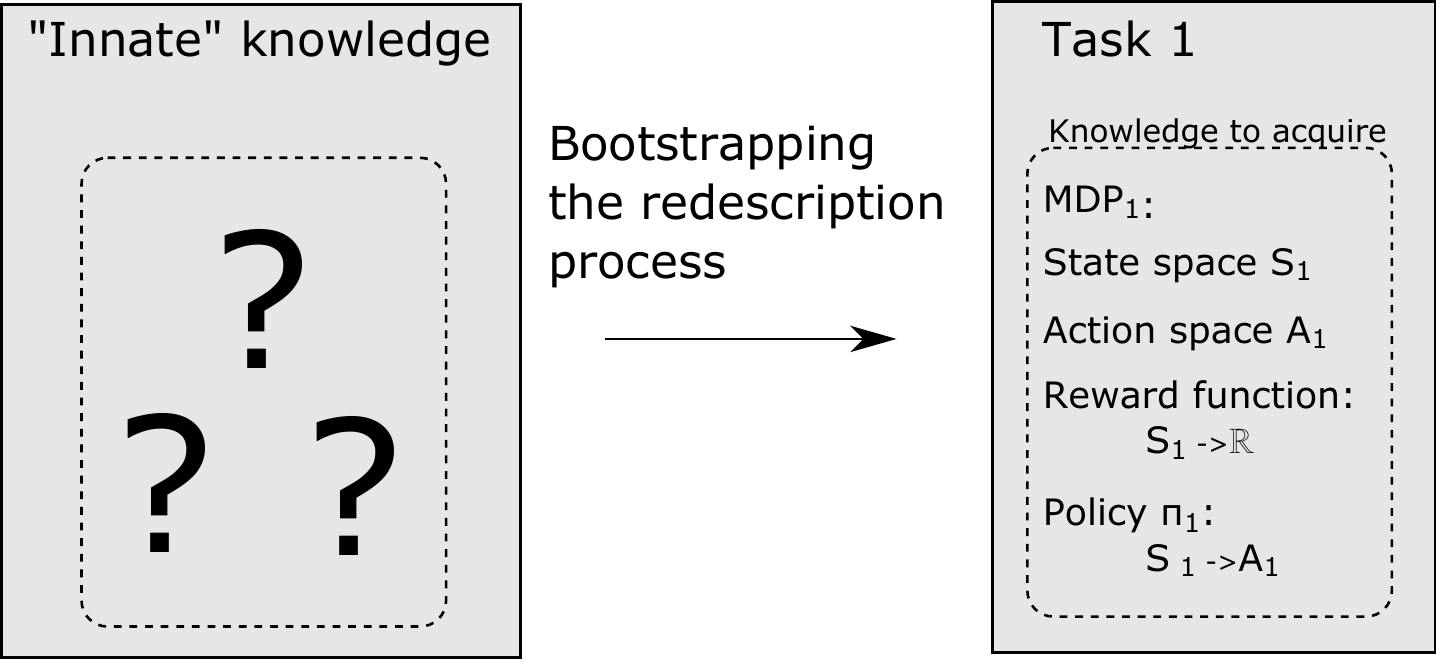}
\end{center}
\caption{\label{fig:bootstrap_scheme}The bootstrap problem: how to generate a first Markov Decision Process when little is known about the task and the domain?}
\end{figure}

\begin{figure}[htb!]
\begin{center}
\includegraphics[width=\linewidth]{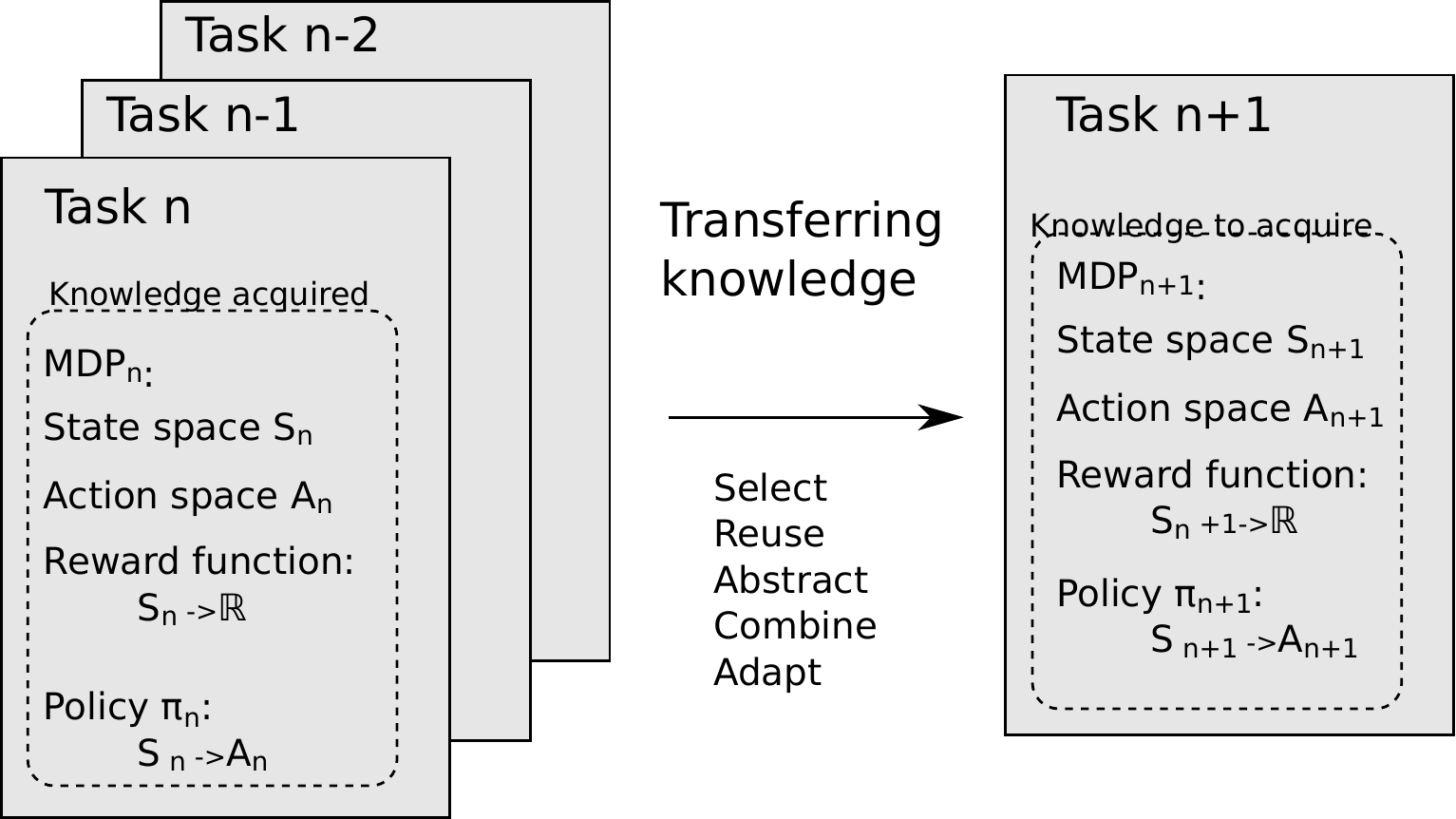}
\end{center}
\caption{\label{fig:transfer_0}The transfer learning problem: how to build a new MDP from a set of known MDPs?}
\end{figure}

In this context, the robot behavior can be described by three loops: (1) Operational, (2) Learning and (3) Redescription (Figure \ref{fig:3loops}). The operational loop corresponds to the execution of a known policy. The learning loop is in charge of acquiring new and appropriate policies from a known representation of states and actions. The redescription loop corresponds to the acquisition of a (more) adapted knowledge representation. This loop is called redescription to emphasize the iterative nature of the process \cite{stoytchev2009some}: the knowledge representation acquisition process necessarily starts with a representation\footnote{The lowest possible level is the raw sensori-motor flow, which is already a representation.} and aims at building a new one. It is thus a transformation more than a creation.

The focus of this work is on the outer redescription loop whose goal can be described as follows: \textit{How to acquire the knowledge required to learn the underlying (state, action) representation of policies able to solve the tasks the robot is facing, when those tasks are not known to the robot designer?} This question raises three different challenges: how to bootstrap the process and build the first state and action spaces when little is known about the task and the domain (Figure \ref{fig:bootstrap_scheme})? How to consolidate the acquired knowledge to make generated policies more robust? and then how to transfer acquired representations to a new task (Figure \ref{fig:transfer_0})?

The article begins with a discussion of the challenges raised by the adaptive capability we are looking for and that we call "open-ended learning", and the representational redescription it implies. The following section presents an overview of DREAM architecture (Deferred Restructuring of Experience in Autonomous Machines), the proposed approach to deal with these challenges developed during the DREAM European project\footnote{\url{http://dream.isir.upmc.fr/}}. The next sections introduce in more details how we have dealt with four of the challenges raised by representational redescription: bootstrapping of the process, state representation acquisition, consolidation of acquired knowledge and knowledge transfer. Next, a related work section follows before a discussion mainly oriented towards the link of this work with Neuroscience.

\section{Open-ended learning}
\label{sec:open-ended_learning}

\subsection{A definition of open-ended learning}

In this work, open-ended learning is an adaptation ability with two major features:
\begin{enumerate}
\item the system can learn a task when both the task and the domain are unknown to the system designer,
\item the system acquires experience along time and can transfer knowledge from a learning session to another. 
\end{enumerate}

The proposed definition is related to two concepts: life-long learning and open-ended evolution. 

Lifelong learning \cite{thrun1998lifelong, parisi2019continual}, also called never ending learning \cite{mitchell2018never} or continual learning \cite{lesort2020continual} consists in going beyond a single learning session and considers that the robot may be faced with different tasks in different environments. The goal is to avoid to start from scratch for each task and exploit acquired knowledge when considering a new task \cite{pan2010survey,torrey2010transfer} while avoiding catastrophic forgetting. In Thrun and Mitchell's original definition, the state and actions spaces are common between the different tasks and known beforehand. Open-ended learning proposes to go beyond this view, considering that using a single action space and a single state space is a strong limitation to the adaptive ability of the system.

Open-ended evolution is a major feature of life \cite{bedau_nature_1996}. It is described as the ability of nature to continuously generate novel \cite{lehman_abandoning_2010,standish2003open} and adapted \cite{bedau_nature_1996} lifeforms. Open-ended evolution considers large timescales and how new lifeforms can emerge as a result of the dynamics of an evolutionary process. It does not consider a single individual, but a species or even a whole ecosystem. We propose to define open-ended learning similarly, but with a focus on a single individual. It could thus be defined as the \textit{ability to continuously generate novel and adapted behaviors}. \textit{Novel} suggests the ability to explore and find new behaviors while \textit{adapted} suggests that these behaviors fulfill a goal.
This association between novelty and adaptation can also be called creativity \cite{doncieux2016creativity}. Another definition of open-ended learning could thus be the \textit{ability to continuously generate creative behaviors}.

 \subsection{Goals and challenges}
 \label{sec:goalsandchallenges}

A robot with an open-ended learning ability is expected to solve all
the tasks it is facing without the need for its designer to provide it
with appropriate state and action spaces \cite{doncieux2018open}. It
implies that the system is not built to solve a single task, but needs
to solve multiple tasks in a life-long learning scenario. We will thus
make the following assumptions: (1) the robot will be confronted to
$n$ different tasks, with $n>1$ and (2) the robot may be confronted
several times to the same task.
In this context, the representational redescription process aims
either at making the robot able to solve a task that was previously
unsolvable (or at least unsolved) or at making its resolution more
efficient when it encounters it again. The goals of the
redescription processes can thus be described as follows:

\begin{itemize}
\item Bootstrapping task resolution: solving a previously unsolved task without task-specific state and action spaces;
\item Improving over experience: increasing efficiency, speed and accuracy of solving a particular task;
\item Generalizing through the transfer of knowledge. Using already acquired
representations to get more \textit{robustness} and \textit{abstraction};
\item Changing the learning or decision process: building the
representations required by a different, and more efficient learning
and decision process in order to move towards zero-shot learning to increase \textit{robustness} and \textit{abstraction}.
\end{itemize}

In this description, \textit{robustness} is defined as the ability to address the same task, but in a different domain, and \textit{abstraction} as the ability to rely on the knowledge acquired while solving a task to address another one.

These goals raise different challenges for representational
redescription processes. Some are shared with learning algorithms
challenges. Dealing with sparse rewards is an example: from a learning
perspective, the challenge is to find a learning algorithm with an
appropriate exploration strategy and from the representational
redescription perspective, the challenge is to find state and action
spaces that increase the probability to succeed by discarding
irrelevant state dimensions, for instance, or by restraining the
actions to those leading to success. The corresponding challenge
can thus be faced either by adapting the learning process or by finding an
appropriate representation.

Some other challenges are specific to representational
redescription. In a reinforcement learning scenario, the state and
action spaces are supposed to be well chosen by the system designer.
Finding them for a robotics setup is notoriously hard
\cite{kober2013reinforcement} and if they are not well designed, it is
expected that the system will not be able to learn an efficient policy
and the fault will be on the system designer. In an open-ended
learning setup, it cannot be assumed that relevant state and action
spaces are initially available and what makes them
relevant needs to be defined. A state space is useless if it does
not provide the system with the information it needs to decide what
action is to be performed. It is also required to interpret an
observed reward. In an MDP, the reward function associates a value to a
state\footnote{The reward function can also be defined on different spaces, for instance on a (state, action) tuple.}. It means that the
system designer determines what reward value results from the system
action, \textit{but also to what state this value is associated}. In a
representational redescription loop, \textit{understanding} an
observed reward value, i.e. finding the state space that best explains
the observation, is a challenge per se.

Finally, \cite{doncieux2018open} have identified eight challenges for
representational redescription. They can be split into two groups: those related to solving a single task and those related to solving multiple tasks:
\begin{itemize}
\item Single task challenges:
\begin{enumerate}
\item Interpreting observed reward: building (or selecting) a state space that makes observed reward predictable and reachable (with an appropriate policy);
\item Skill acquisition: building the actions to control the state space;
\item Simultaneous acquisition of state and action spaces as well as policies;
\item Dealing with sparse rewards, in particular when bootstrapping the redescription process;
\end{enumerate}
\item Multi-task challenges:
\begin{enumerate}
\item Detecting task change;
\item Ordering knowledge acquisition and task resolution;
\item Identifying the available knowledge to build a new MDP;
\item Transferring acquired knowledge.
\end{enumerate}
\end{itemize}

\section{Overview of the proposed approach}

We now present the DREAM approach to deal with some of the challenges identified in the previous section. The approach is focused on the acquisition of knowledge through interactions of the robot with its environment and follows a stage-by-stage developmental process, where some stages rely on an evolutionary approach. It is thus an \textit{Evolutionary Developmental Intelligence} approach to Artificial General Intelligence \cite{doya2019toward}. This section describes its main features and the following sections describe the implementation done so far and the results we have obtained. 

\subsection{Asymptotically end-to-end}

One of the main limitations of robotics that has motivated this work is the lack of flexibility of a system limited by a single predefined representation. On a single task, carefully designing the state space, the action space and choosing an appropriate policy representation may lead to impressive results \cite{kober2013reinforcement}, but changing the task or the domain requires a new design phase and thus reduces the robot adaptivity. To maximize the robot versatility, it should be able to rely on the lowest possible level, for both the sensory and motor information: its learning process should be able to exploit the raw sensori-motor data. These approaches are named end-to-end \cite{levine2016end} to highlight this capacity to start from the very first data entering the system and generate the data expected from its motors. Any intermediate representation may make learning and decision easier, but it is defined with some a priori in mind that may fit well to some tasks but not to others. 

For the perception part, and with a focus on vision, the a priori may be on the kind of relevant information: is it static information (shape, color) or dynamic information (motion)? Does it involve large areas (walls), or small ones (pens)? Does it have a homogeneous texture, or is it made up with parts having different features? Are there "objects"? And if it is the case, are they solid or deformable? Many other questions of this kind can be raised that will influence the perception model. And the robot won't be able to deal with a new situation that requires some perceptions that have not been covered by the models implemented in the system.

It is the same for the action part: what is important in the robot motion? Is it a question of position control? Velocity control? Torque control? Is it open-loop, closed-loop? If closed loop, what information needs to be taken into account to adapt robot trajectory? As for the perception part, any choice made at this point will limit the final adaptivity of the robot.

End-to-end approaches require to use a learning method that can deal with high dimensions, both as input and as output. Deep learning is the only approach so far that has been able to deal with end-to-end control \cite{levine2016end, james2017transferring,pan2018agile}. Neural networks can deal with these large spaces but at a condition: a large enough training database must exist. It raises a critical bootstrap challenge: how to collect enough \textit{meaningful} data to train the system? A set of random motions will likely not be appropriate. An arm robot randomly moving, for instance, will only very rarely interact with objects \cite{maestre2015bootstrapping}. There is then little chance that the features extracted by the deep neural networks describe them with enough accuracy, thus impeding the convergence of any learning or decision process on an object interaction task.

Several approaches have been used to generate the required data. Some rely on a simulation \cite{james2017transferring}, but require the 3D structure of the environment. Others rely on demonstrations \cite{pan2018agile}, but providing such demonstrations is not straightforward in an open-ended learning scenario. Other approaches reduce the number of required samples by carefully defining the cost function and adapting it to the task with a fitting phase that require human intervention \cite{levine2015learning}. All these approaches show that end-to-end learning is possible, but also highlight the challenge of acquiring relevant data in an open-ended learning scenario. 

The necessary information, may it be demonstrations, the 3D structure of the environment or dedicated cost functions, could actually be acquired in a preliminary stage. We propose to add some processes that rely on predefined representations in order to bootstrap the system and acquire these data. The difference with other approaches like options \cite{sutton1999between}, is that these representations are not a basis on which the whole system is built, as new and independent representations relying on the raw sensori-motor flow can be acquired. After a while, predefined representations may not be required anymore. This is why we have called this feature \textit{asymptotically end-to-end}: the system starts with predefined representations and once it has acquired enough experience, it can start building, from the raw sensori-motor flow, new representations that future learning and decision processes can rely on. 

\subsection{Focus on representational redescription}

The end-to-end approaches evoked so far rely on a single neural network architecture that goes directly from the raw sensors to the raw effectors without any intermediate step. This is an advantage as it reduces the engineering effort related to the definition of the corresponding architecture and it reduces the biases due to the designer choices. 

We have made a different choice and we put the focus on the internal representations that are built by the system. Instead of considering them as an internal and somewhat hidden information, that is a by-product of the end-to-end learning process, we explicitly look for those representations, in the form of MDPs. The goal is to build algorithms that create such representations with the features required by the available learning or decision algorithms (is the representation continuous or discrete, in large or small dimensions, etc.). Aside from enabling the use of existing approaches \cite{jonschkowski2015learning,Lesort18}, the goal is also to make the system more transparent: an analysis of these representations directly tells what the robot perceives and what it can or cannot do. Another advantage is that it allows us to decompose the problem and define processes focused on state space acquisition and others on action space acquisition. It also separates the open-ended learning process into two different phases (Figure \ref{fig:acquis-exploit}): knowledge acquisition (building new representations, but also acquiring the experience required by this process) and knowledge exploitation (task resolution exploiting the representations found so far).

\subsection{Stage by stage modular approach}

As long as a single process cannot fulfill all the requirements of open-ended learning, it is interesting to decompose the process and clearly identify the inputs and outputs of each part in order to develop them in parallel. This basic software engineering consideration has lead us to decompose the core of our approach, i.e. representational redescription, as a set of modules, each having a clear input and a clear output. The inputs are the required knowledge for the module to be used and the output is the knowledge it builds. Each module can thus be connected to other modules, through a knowledge producer and customer link, resulting in a graph of dependencies. Under this view, an open-ended learning process can be described as a graph of interconnected modules with the first modules that require limited knowledge about the task and the environment, and the last modules that result in task-specific representations (e.g. an MDP) that a learning or decision process can exploit to solve the current task. A cognitive architecture can then select the modules to activate at a given instant in time given their constraints and what the system currently aims at. The implementation described later is a first proof-of-concept that does not include this latter module selection part.

This approach has another consequence: it makes it possible to rely on any kind of learning algorithm, as long as it is possible to build a module or a chain of modules that builds the required knowledge. 

\begin{figure}[htb!]
\begin{center}
\includegraphics[width=\linewidth]{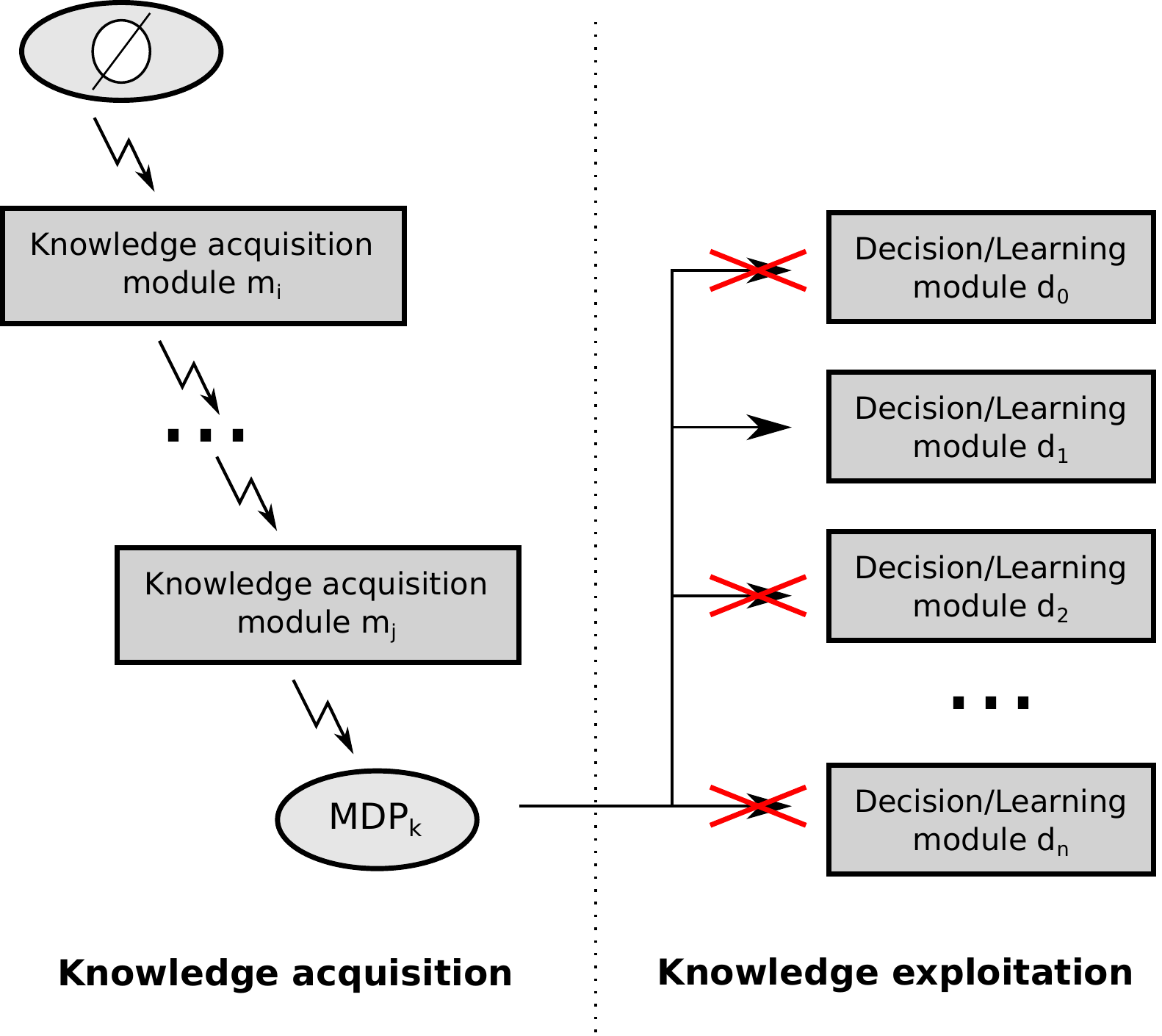}    
\end{center}
\caption{\label{fig:acquis-exploit}Example of knowledge acquisition chain going from no knowledge about a task $k$ ($\emptyset$) to the design of a dedicated MDP ($MDP_k$). This MDP is compatible with the decision or learning process $d_1$, that can exploit it to solve the task.}
\end{figure}

\begin{figure*}[htb!]
\begin{center}
\includegraphics[width=\linewidth]{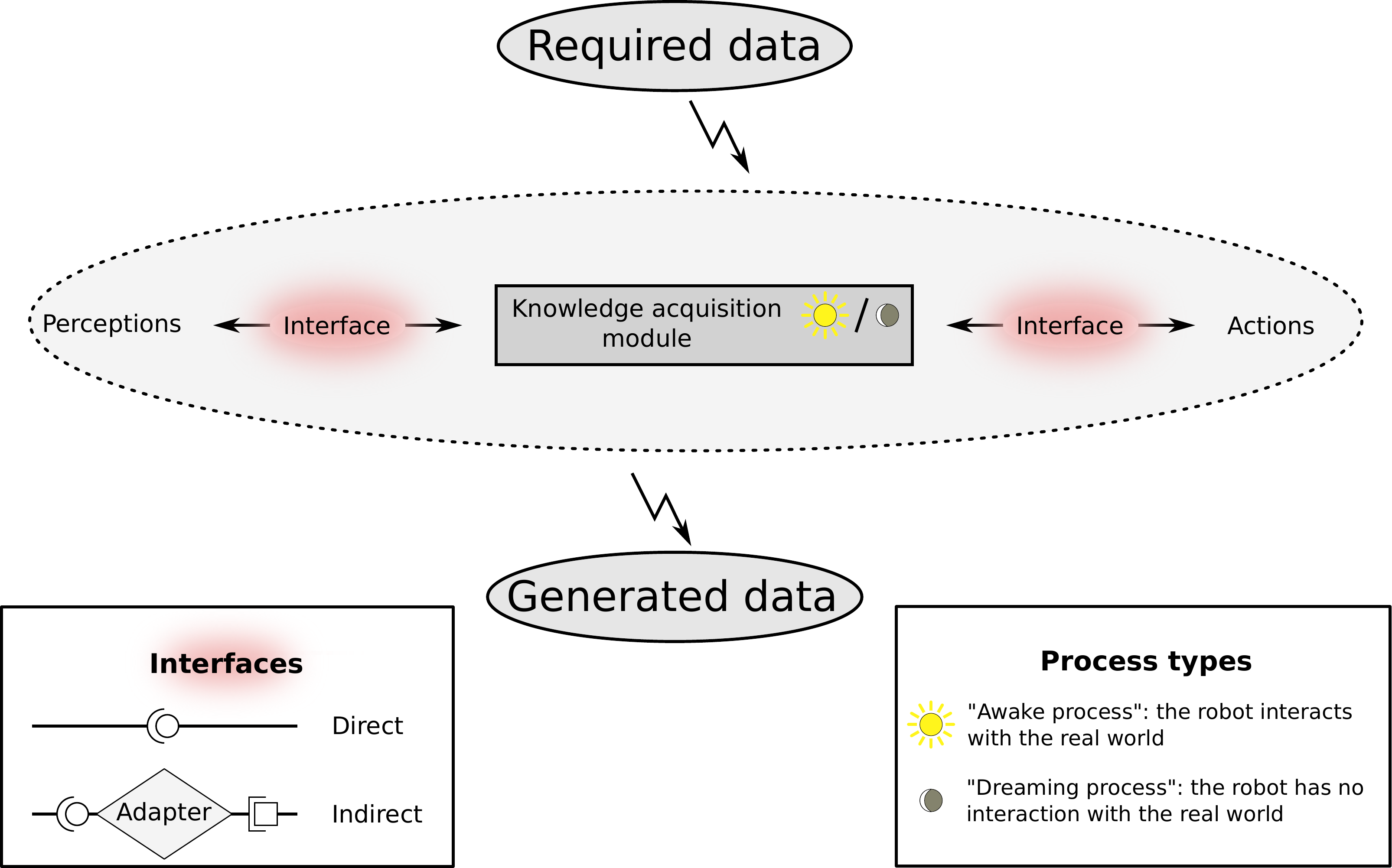}    
\end{center}
\caption{\label{fig:dev_modules}Zoom in on a developmental module. A developmental module has required data and generates data. It can directly use the raw sensorimotor flow or it may require an adapter (a perception model or a motor primitive, for instance). The connections between the module and the perceptions and actions are in both directions as some modules read perception values and generate actions, but others may generate new perceptions (data augmentation methods) or read raw actions (action restructuring modules). Finally, a module may require the robot to perform new interaction with the environment ("awake process") or not ("dreaming process"). This distinction is important as "awake modules" create specific constraints (they need access to the robot, they may damage the robot or the environment, and they have a significant cost in terms of time and mechanical wear and tear). Dreaming processes can be executed in parallel. They typically rely on previously acquired sensori-motor data, may it be directly, or through models learned or tuned out of these data.}
\end{figure*}

Figure \ref{fig:acquis-exploit} illustrates this modular approach. Figure \ref{fig:dev_modules} shows a single module to highlight its features. To be end-to-end, the chain building the MDP requires to have at least one module directly connected to the raw sensor flow and at least one module (that may be the same) directly connected to the motor flow. These modules need to be in charge of building resp. the perception part (state space) and the motor part (action space) for learning and decision processes.

\subsection{Alternation between awake and dreaming processes}

Most animals alternate between awake and sleeping phases. While the awake phases are clearly important for the animal to survive, the role of the sleeping phases has been revealed only recently, at least concerning its impact on cognition. These phases are notably related to memory reprocessing \cite{stickgold2001sleep}. They may involve the replay of past events to consolidate learning \cite{de2015explicit,caze2018hippocampal} or the exploration of new problem solving strategies \cite{wagner2004sleep,gupta2010,khamassi2020}. We think that this distinction is also important in robotics and, as shown in Figure \ref{fig:dev_modules}, we propose to highlight this module category. These two kinds of modules will not impose the same constraints. Awake modules require interactions with the real world. They need to control the robot and may produce damages or at least increase the robot's wear and tear. Each interaction has then a high cost. "Dreaming" processes do not create such constraints. They may correspond to data analysis or exploration through the means of a model of the world and their sampling cost is then significantly lower. They should then be preferred.



\subsection{Development}

\begin{figure*}[htb!]
\begin{center}
\includegraphics[width=\linewidth]{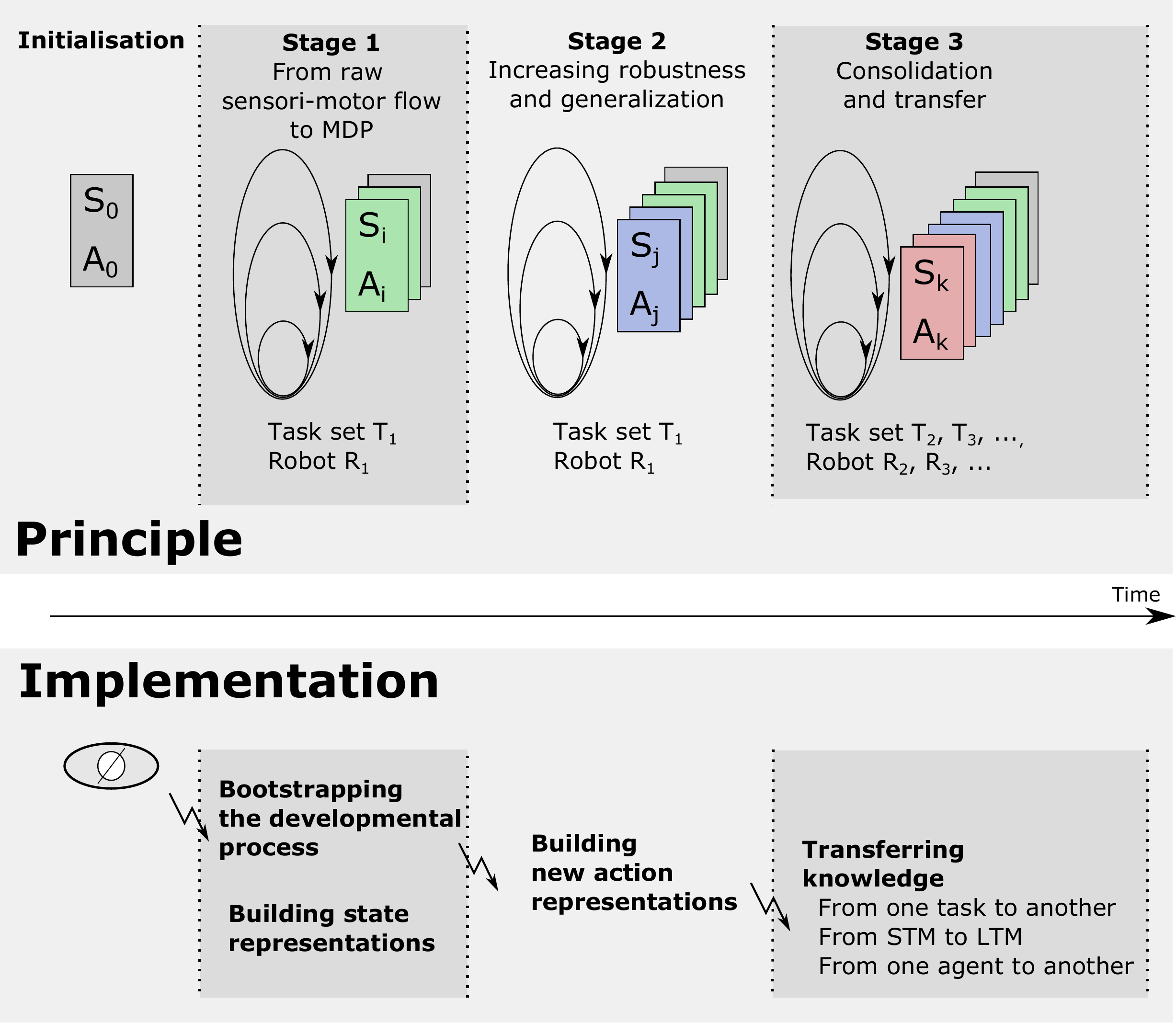}
\end{center}
\caption{\label{fig:overview}Overview of the DREAM approach. Starting from raw sensorimotor values, stage 1 processes bootstraps the process and builds a first set of representations. Stage 2 processes consolidates these representations on the same set of tasks. Stage 3 processes further restructure representations and acquire the knowledge required to facilitate transfer between tasks, allow knowledge reuse and sharing between robots. $S_n$ is the $n$-th state space and $A_n$ the $n$-th action space. The lower part of the figure indicates what has been implemented so far and gives the name of the corresponding sections.}
\end{figure*}

The knowledge acquisition chain can be split up into different stages, each stage resulting in an MDP that can, to some extent, solve a task. By reference to biology and child development, we propose to call this approach \textit{development}, as we aim to reproduce some of the functions of human development. It highlights that the proposed open-ended learning process does not homogeneously drive the robot from its naive performance at startup to its acquired expertise after enough learning and practice. As in Piaget's models of development \cite{piaget2003part}, the proposed open-ended learning follows a succession of stages that have their own features and associated modules. Each stage corresponds to one or several knowledge acquisition modules and can be described by a particular goal, stage $n$ being necessary to execute stage $n+1$ (Figure \ref{fig:overview}). As in the overlapping wave theory \cite{siegler1998emerging}, stage $n$ is not supposed to be strictly stopped before stage $n+1$ is activated. Stage $n$ can be reactivated after stage $n+1$ has been started. Anyways, this point goes beyond the scope of this article.

It should be noted that we do not aim at building a model of child development, but that our goal is similar for robots to what psychologists and neuroscientists attribute to child development. Put differently, Psychology and Neuroscience may be a source of inspiration, but not a constraint and elements of the current implementation do not systematically have a Neuroscience counterpart. Building such a system can anyway lead to new insights in the neuroscientific study of related processes. This point is further discussed in the Section \ref{sec:disc_neuro}. 

The following sections describe the current implementation of the proposed approach. Figure~\ref{fig:overview} puts each section in the perspective of the whole proposed developmental scheme.

\section{Building state representations}
\label{sec:state_repr}


\begin{figure}[htb!]
\begin{center}
    \includegraphics[width=0.8\linewidth]{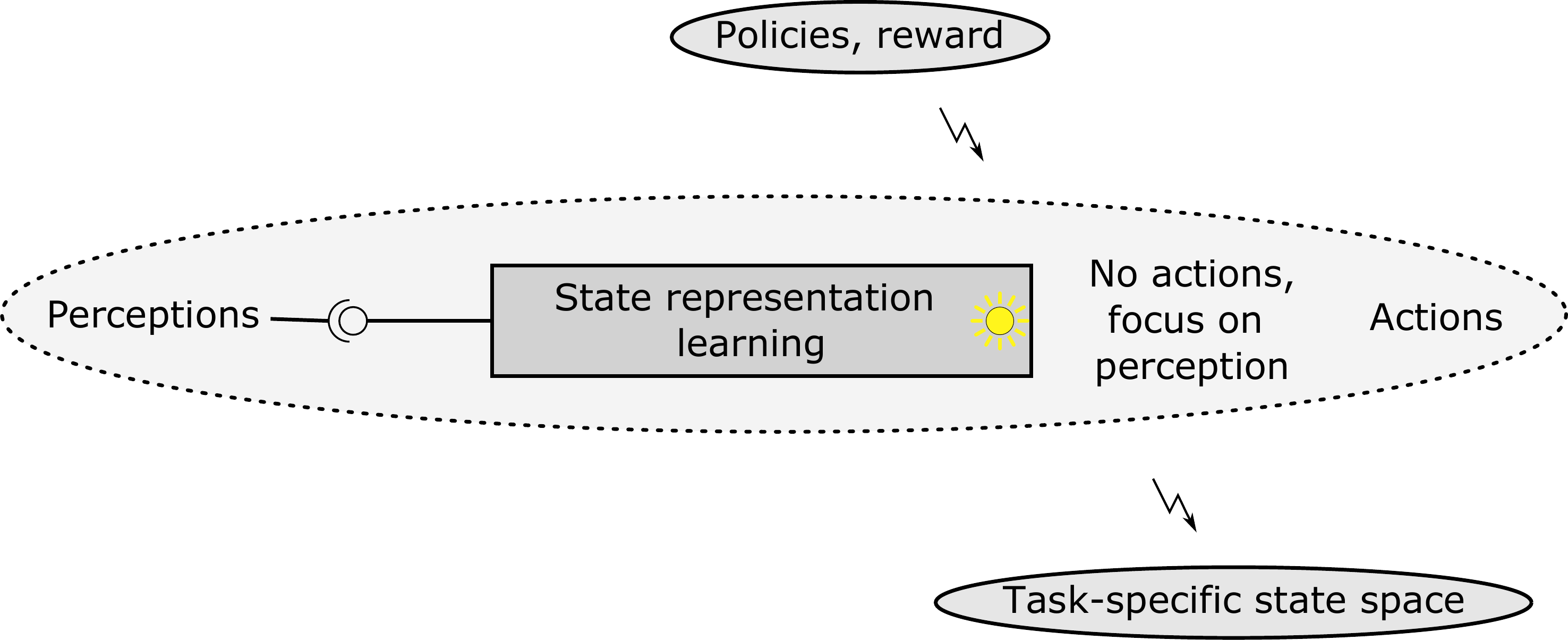}
\end{center}
\caption{\label{fig:srl_step}State representation learning module. Starting from a reward that defines a task and policies to observe at least some rewards, it generates a task-specific state space. The approach relies on raw perceptions. Actions are performed by the provided policies and during the validation step, once the state space has been generated. No action is proposed by the state representation module.}
\end{figure}

State Representation Learning (SRL) is the process of learning, without explicit supervision, a representation extracted from the observations that is adapted to support policy learning for a robot on a particular task or set of tasks. States are the basis of MDPs. They contain the required information to make a decision. Making the right decision to reach a goal implies some exploration of this state space that consequently needs to be low-dimensional for the planning or learning to be efficient. At the same time, the acquisition of a new state space requires to generate observations that cover what the robot may experience while solving a task. The design of a state space is then a chicken-and-egg problem as a policy is required to generate observations to be used later on to generate a new and relevant state space that can be used to learn policies. This problem has been tackled here with random policies on a simple button pushing task and the next section shows how more complex policies could be generated to bootstrap the generation of state spaces for more complex tasks.

 Our state-of-the-art survey \cite{Lesort18} analyzes existing SRL strategies in robotics control that exploit 4 main learning objectives: reconstructing the observations, learning a forward model, learning an inverse model, or exploiting high-level prior knowledge. Methods were also proposed to exploit several of these objectives simultaneously.
Furthermore, we developed and open sourced\footnote{\url{https://github.com/araffin/robotics-rl-srl}} the S-RL Toolbox \cite{Raffin18} containing baseline algorithms, data generating environments, metrics and visualization tools for assessing SRL methods.





We propose a new approach to SRL for goal-based robotics tasks that consists in learning a state representation that is split into several parts where each part optimizes a fraction of the objectives. In order to encode both target and robot positions, auto-encoders, reward and inverse model losses are used: 

\begin{itemize}
\item \textit{Inverse model:} One important aspect to encode for reinforcement learning is the state of the controlled agent. In the context of goal-based robotics tasks, it could correspond to the position of different parts of the robot, or to the position of the tip of a tool hold by the robot. A simple method consists in using an \textit{inverse dynamics objective}: given the current $s_t$ and next state $s_{t+1}$, the task is to predict the taken action $a_t$. The type of dynamics learned is constrained by the network architecture. For instance, using a linear model imposes linear dynamics. The learned state representation encodes only controllable elements of the environment. Here, the robot is part of them. However, the features extracted by an inverse model are not always \textit{sufficient}: in our case, they do not encode the position of the target since the agent cannot act on it.

 \item \textit{Auto-encoder:} The second important aspect is the goal position. Based on their reconstruction objective, auto-encoders compress all information in their latent space, but they tend to encode only aspects of the environment that are salient in the input \cite{Lesort19}. This means they are not task-specific: relevant elements for a task can be ignored and distractors (unnecessary information) can be encoded into the state representation. In our case however, among other information, they will encode the goal position. Therefore, they usually need more dimensions than apparently required to encode a scene (e.g. in our experiments, it requires more than 10 dimensions to encode properly a 2D goal position).

\item \textit{Reward prediction:} The objective of a reward prediction module leads to state representations that are specialized in a task, thus improving the representation of the goal position in our case. Note that without the complementary learning objectives, predicting reward would only produce a classifier detecting when the robot is at the goal, thus not providing any particular structure or disentanglement to the state space.
\end{itemize}

\begin{figure}[h!]
\begin{center}
  \includegraphics[width=0.45\textwidth]{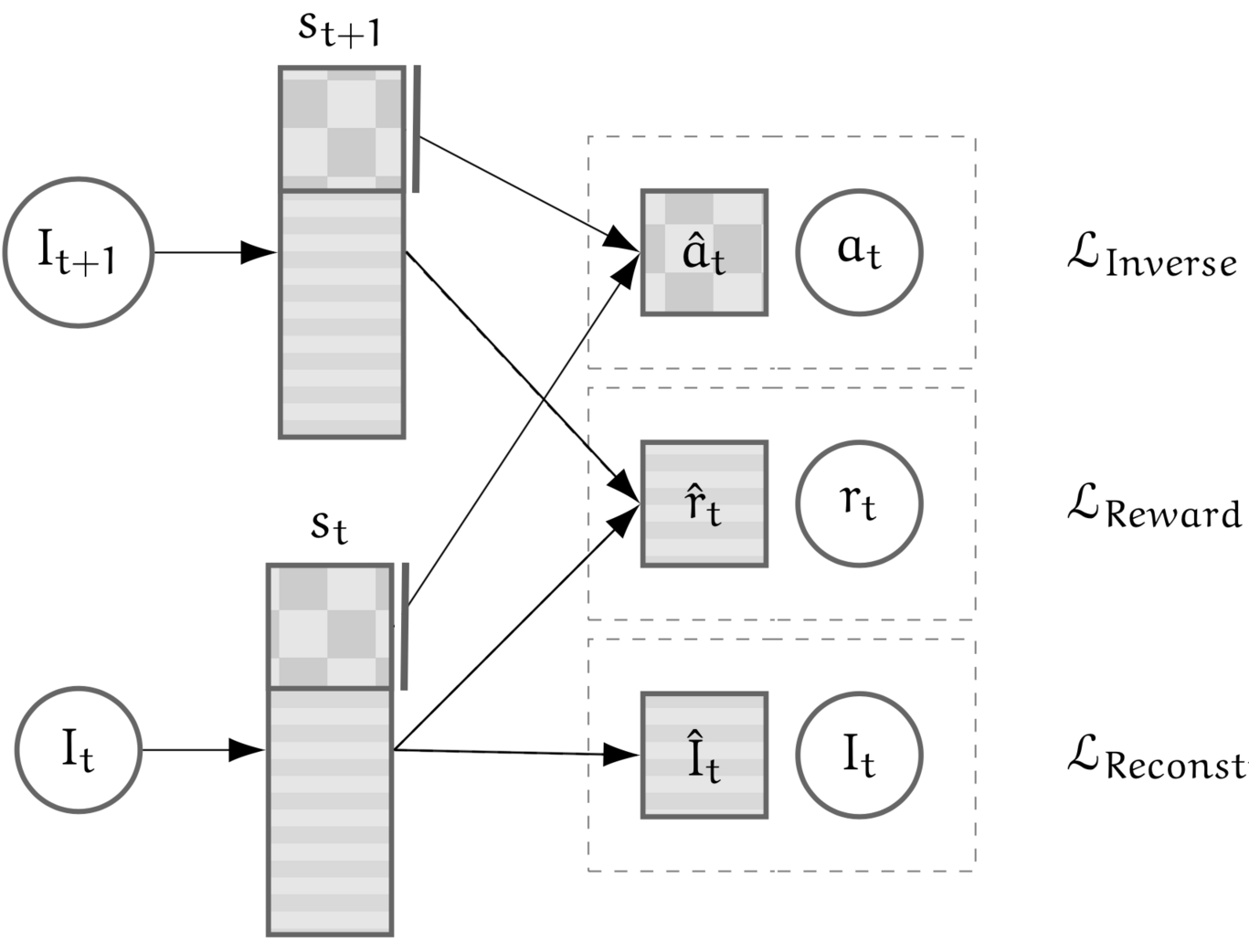}
\end{center}
\caption{\textit{SRL Splits} model: combines the losses of reconstruction of an image $I$ (auto-encoder) and of reward ($r$) prediction on one split of the state representation $s$, and the loss of an inverse dynamics model on the second split of $s$. 
Arrows represent model learning and inference, dashed frames represent losses computation, rectangles are state representations, circles are real observed data, and squares are model predictions \cite{Raffin19}.
\label{fig:split-model}}  
\end{figure}

Combining these objectives into a single loss function on the latent space can lead to features that are \textit{sufficient} to solve the task. However, these objectives are partially contradictory and stacking partial state representations will therefore favor \textit{disentanglement} and prevent opposed objectives from cancelling out, thus allowing a more stable optimization and better final performance.  Fig.~\ref{fig:split-model} shows our split model where each loss is only applied to part of the state representation (see \cite{Raffin19} for a more detailed presentation). 

\begin{figure}[htbp!]
\centering
  \includegraphics[width=0.95\linewidth]{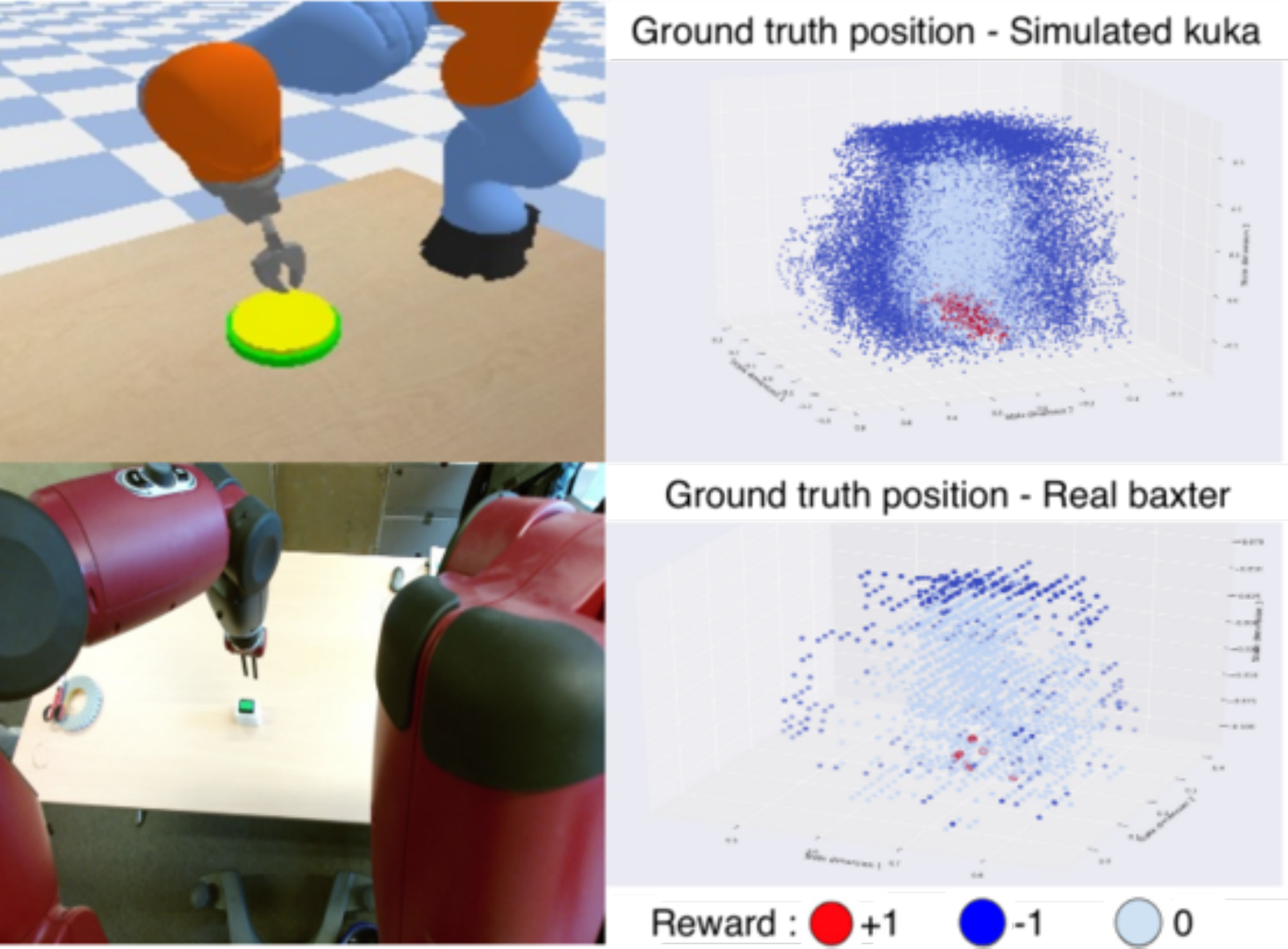} 
  \caption{Top Left: Simulated 3D robotic arm in a random target environment. Top Right: Ground truth gripper positions and associated reward (-1: arm out of sight; 1: gripper touching button; 0: otherwise).
  Bottom Left and Right: similarly for Baxter real robot environment with a fixed target.}
  \label{fig:setUp-staticButtonSimplest-ground-truth} 
\end{figure}

We applied this approach to the environments visualized in Fig. \ref{fig:setUp-staticButtonSimplest-ground-truth}, where a simulated arm and a real Baxter robot are in front of a table and image sequences taken from the robot's head camera contain a front view of what the robot is able to see.  We consider a "reaching" (\textit{pushing button}) task with a randomly placed button on the table. In this environment RGB images are 224x224 pixels; 
three rewards are recorded: 0 when the gripper is not touching the button, 1 when touching it, and -1 when the robot gripper is out of the field of view of the frame. The goal of this task is to learn a representation consistent with the actual robot's hand position and button position. Actions are defined by elementary movements of the hand along the X, Y, Z axes in the operational space between timesteps $t$ and $t+1$.

Table \ref{tab:gt-correlation-kuka-3D-arm-random} shows that our approach outperforms several baselines in terms of correlation of learned states with the Ground Truth Correlation (GTC, see \cite{Raffin18}) on the simulated robot, and that using this representation for reinforcement learning using the Proximal Policy Optimization (PPO) algorithm \cite{schulman2017proximal}  performs very close to using the ground truth. Table~\ref{tab:gt-correlation-Baxter-3D-arm-fixed} shows that similar results are obtained for SRL models (GTC) on real robot data.



\begin{table} [ht!]
 \centering 
         
    \resizebox{\columnwidth}{!}{
    \begin{tabular}{ l|lllll|l|l}\hline 
    \textbf{GTC} & \textit{$x_{rob}$} & \textit{$y_{rob}$} & \textit{$z_{rob}$} & \textit{$x_{targ}$}  & \textit{$y_{targ}$} & \textit{Mean} & \textit{Reward} \\\hline
    Ground Truth  &  1 &  1 & 1 & 1 & 1 & 1 &  4.92 $\pm$ 0.10 \\
    Supervised & 0.57 & 0.74 & 1 & 0.79 & 0.69 &  0.76 & 4.89  $\pm$ 0.11\\
    \hline
    Raw Pixels & NA & NA & NA & NA & NA & NA & 4.78  $\pm$ 0.15  \\
    \hline
    Rand. Features & 0.36 & 0.54 & 0.49 & 0.73 & 0.83 & 0.59 & 2.17  $\pm$ 0.44  \\
    Auto-Encoder & 0.43 & 0.73 & 0.67 & 0.57 & 0.50 & 0.58 & 4.84 $\pm$ 0.14 \\
    Robotic Priors & 0.18 & 0.03 & 0.18 & 0.75 & 0.42 & 0.31 & 2.22  $\pm$ 0.43 \\
    \hline
    SRL Splits & 0.83  & 0.87 & 0.72 & 0.53 & 0.63 &  0.72 & 4.90 $\pm$ 0.14  \\
    \hline 
    \end{tabular}
    }
\caption{Ground truth correlation and mean reward performance in RL (using PPO) per episode after 3 millions steps, with standard error (SE) for each SRL method in 3D simulated robotic arm with a random target environment.}
\label{tab:gt-correlation-kuka-3D-arm-random}
\end{table}

\begin{table} [ht!]
 \centering 
         
    \begin{tabular}{ l|lll|l}\hline 
    \textbf{GTC} & \textit{$x_{rob}$} & \textit{$y_{rob}$} & \textit{$z_{rob}$} & \textit{Mean} \\\hline
    Ground Truth  &  1 &  1 & 1 & 1 \\
    Supervised & 0.99 & 0.99 & 0.99 & 0.99 \\
    \hline
    Random Features & 0.49 & 0.51 & 0.54 & 0.51\\
    Auto-Encoder & 0.63 & 0.77 & 0.67 &  0.69 \\
    Priors & 0.37 & 0.25 & 0.79 & 0.47 \\
    \hline
    SRL Splits & 0.89 & 0.88 & 0.69 & 0.82 \\
    \hline 
    \end{tabular}
\caption{Ground truth correlation (GTC) for each SRL method in real Baxter robotic arm with a fixed target environment.}
\label{tab:gt-correlation-Baxter-3D-arm-fixed}
\end{table}

\begin{figure}[ht!]
 \centering 
 \includegraphics[width=\columnwidth]{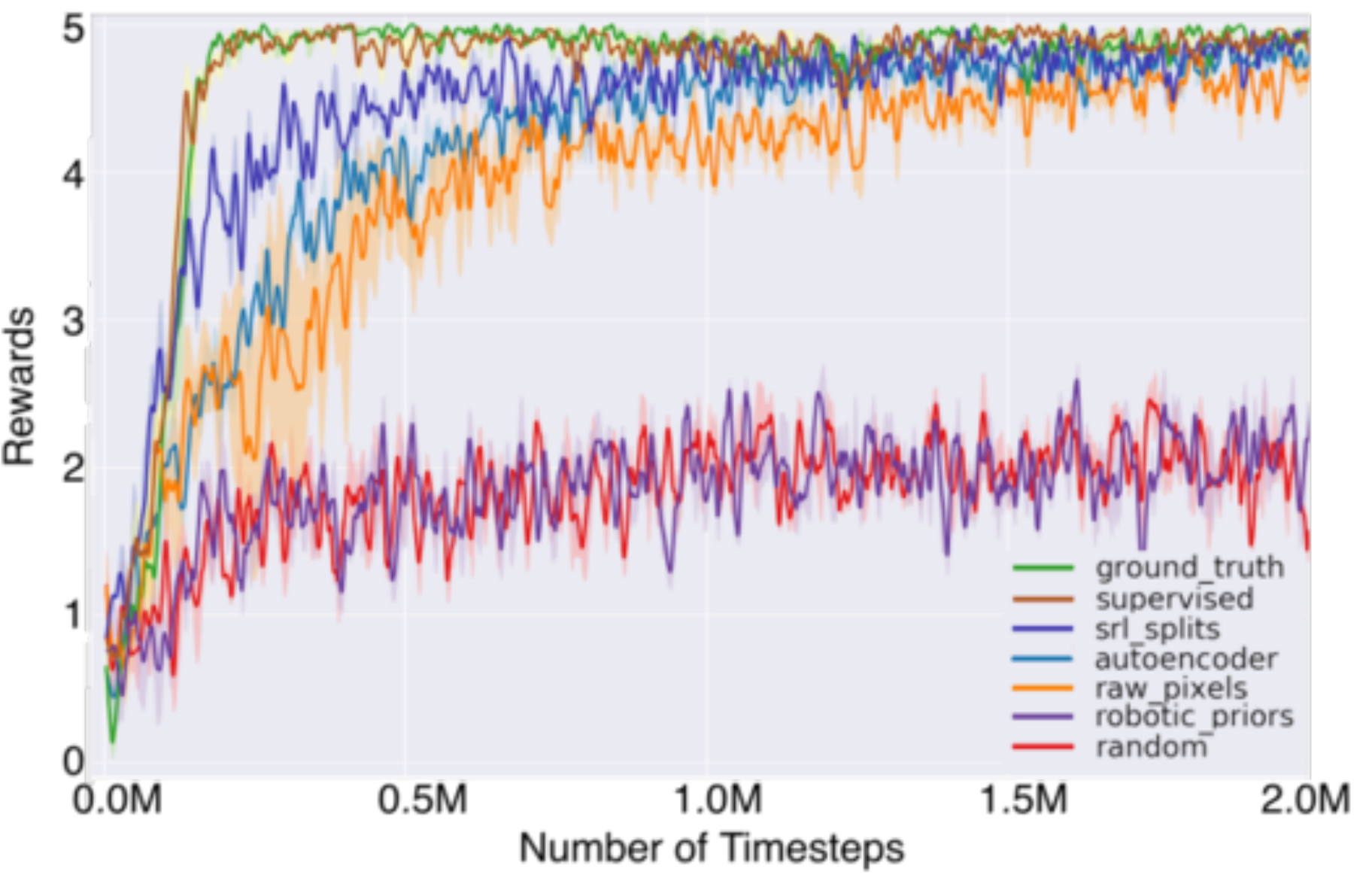} 
 \caption{Performance (mean and standard error for 3 runs) for PPO algorithm for different state representations learned in the 3D simulated robotic arm with random target environment.}
 \label{fig:ppo2-kuka-3d-random-target}
\end{figure}


Learning curves in \figurename~\ref{fig:ppo2-kuka-3d-random-target} show that with our proposed approach, RL can use this learned representation to converge faster towards the optimal performance than with unsupervised alternatives, and at a speed close to the one of the ground truth and the supervised learning cases. The newly defined MDP, based on the learned states, is therefore better adapted to solve this task than the original MDP based on raw images. This new MDP could then be exploited for other tasks that require the same information. 

This experiment shows that a task-specific state space can be learned from raw observations. It requires to define a policy or several policies that will generate the data necessary to learn the perception-to-state mapping. This is an important issue as the generated representation critically depends on the observed data. In the experiments reported here, a random policy has been used. Scaling this approach to more complex tasks requires to be able to generate more advanced policies. The next section introduces the proposed approach to bootstrap this knowledge acquisition system. It allows the robot to learn the structure of the environment, i.e. the objects it can interact with and open-loop policies that can be used later on to learn new states with the current representation. This knowledge paves the way to the acquisition of new state representations in which rewards are self-built on the basis of expected effects on identified objects and the policies to generate the required data are the open-loop policies generated by the proposed learning approach. 


\section{Bootstrapping the developmental process}
\label{sec:bootstrap}

\begin{figure*}[htb!]
\begin{center}
    \includegraphics[width=0.6\linewidth]{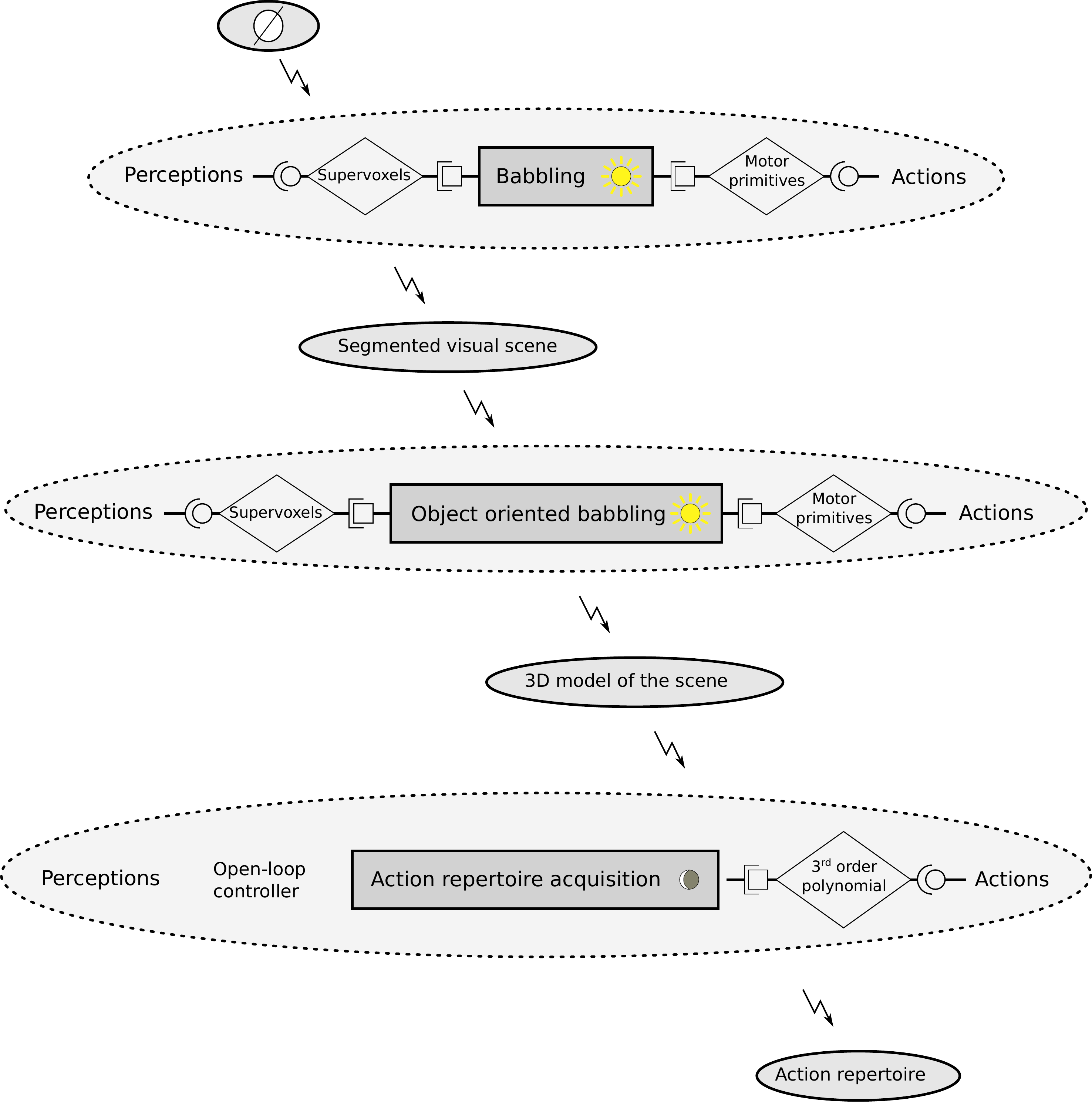}
\end{center}
\caption{\label{fig:bootstrap}The three modules of the bootstrap phase. The result is a repertoire of actions which can be used  either to solve simple problems if the state space is known, or as a training set, or else to collect sensori-motor data.}
\end{figure*}

The bootstrap phase is critical to collect enough data for the later representation redescription modules. It has been split into three modules (Figure~\ref{fig:bootstrap}) that result in a repertoire of actions that can either be used directly to solve simple tasks or as a training set for learning new state spaces (Section \ref{sec:state_repr}) or action spaces (Section \ref{sec:action_repr}). These modules deal with the following challenges:
\begin{itemize}
    \item Skill acquisition: building the actions to control state spaces identified so far and pave the way to the acquisition of new and more relevant state spaces;
    \item Dealing with sparse rewards, in particular when bootstrapping the redescription process;
\end{itemize}

The bootstrap phase implies learning states and actions. But it relies on predefined representations of these spaces to bootstrap the system and acquire the data required by the redescription processes presented in the other sections.

\begin{figure*}[h!]
\centering
\subfloat[Relevance map after 1 interaction]{\label{fig:rm_it1}
\includegraphics[width=.19\linewidth]{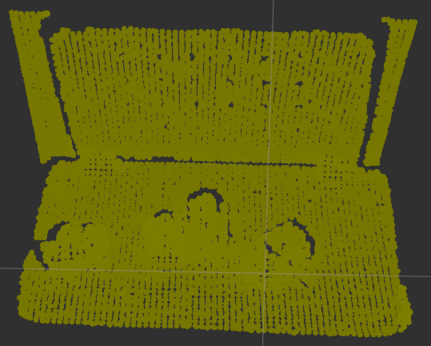}
} 
\subfloat[Relevance map after 10 interactions]{\label{fig:rm_it10}
\includegraphics[width=.19\linewidth]{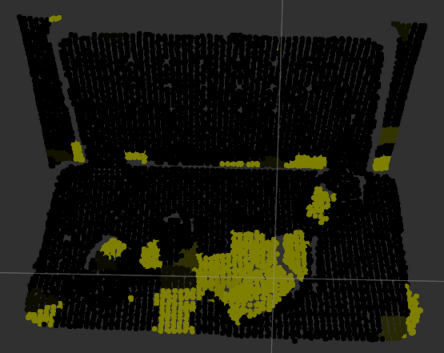}
}
\subfloat[Relevance map after 50 interactions]{\label{fig:rm_it50}
\includegraphics[width=.19\linewidth]{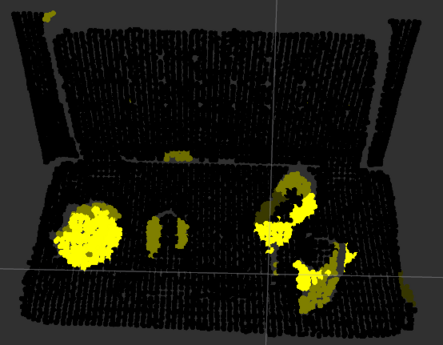}
} 
\subfloat[Relevance map after 100 interactions]{\label{fig:rm_it100}
\includegraphics[width=.19\linewidth]{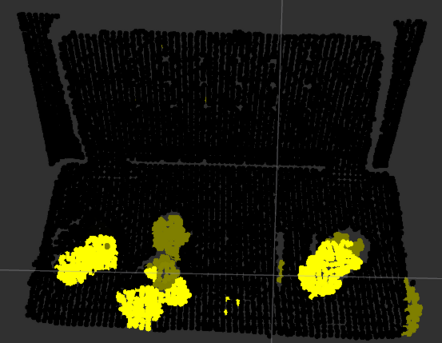}
} 
\subfloat[Relevance map after 400 interactions]{\label{fig:rm_it400}
\includegraphics[width=.19\linewidth]{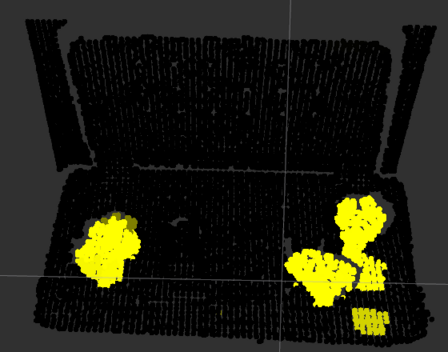}
} 
\subfloat{
\includegraphics[width=.013\linewidth]{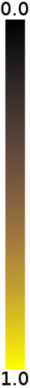}
} \\
\subfloat[Pointcloud used to generate above image]{\label{fig:cloud_it1}
\includegraphics[width=.19\linewidth]{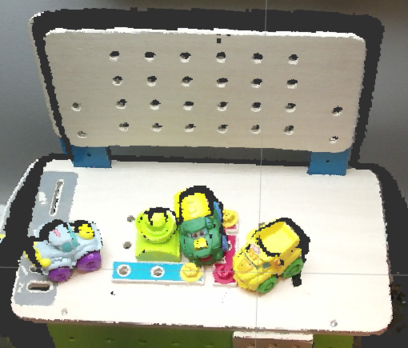}
} 
\subfloat[Pointcloud used to generate above image]{\label{fig:cloud_it10}
\includegraphics[width=.19\linewidth]{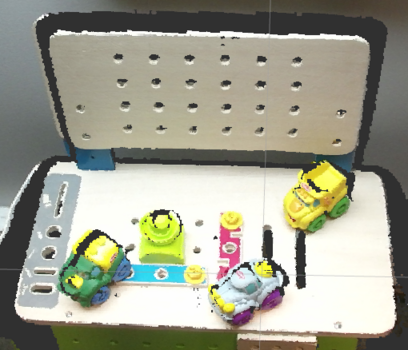}
}
\subfloat[Pointcloud used to generate above image]{\label{fig:cloud_it50}
\includegraphics[width=.19\linewidth]{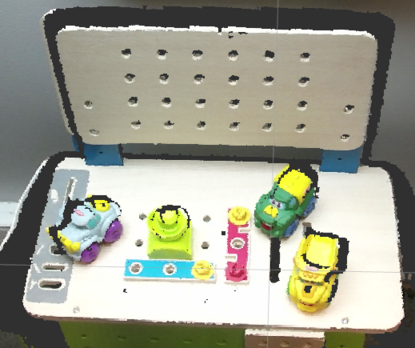}
} 
\subfloat[Pointcloud used to generate above image]{\label{fig:cloud_it100}
\includegraphics[width=.19\linewidth]{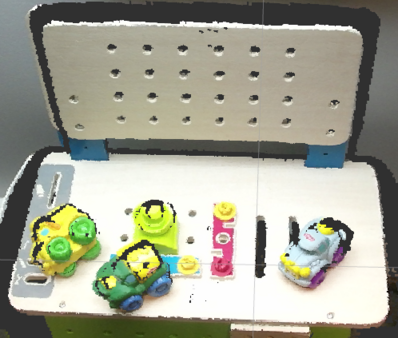}
} 
\subfloat[Pointcloud used to generate above image]{\label{fig:cloud_it400}
\includegraphics[width=.19\linewidth]{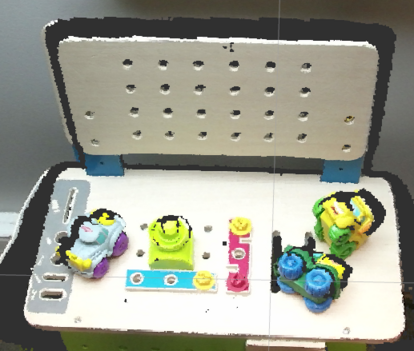}
} 
\caption{Sequence of pointclouds representing a relevance map at different points during exploration. These images have been generated after exploration. Adapted from \cite{goff2019bootstrapping}.}
\label{fig:rm_seq}
\end{figure*}

\subsection{Babbling to identify objects}

To identify objects, a two-step approach is proposed. In a first step, a segmentation separating the background from the parts with which the robot can interact is learned. Then, from this segmentation, 3D object models are learned. Both steps are based on the interactive perception paradigm \cite{bohg2017interactive}. The robot explores an environment by interacting with it in order to collect data and train models on them. This interaction relies on predefined motor primitives. Once the structure of the environment has been identified, new motor primitives are learned that can replace the ones provided at startup, thus implementing the asymptotic end-to-end principle.

This two-step approach relies on minimal environment-specific assumptions. Indeed, the first step builds a simple representation of the environment which does not need a lot of prior knowledge. Then, in the second step, the prior knowledge needed to build object models, like the number of objects or their approximate position, can be easily inferred from the first segmentation.

\paragraph{Relevance Map: A First Segmentation of the Environment.}

In the first step, the robot builds a perceptual map called \textit{relevance map} by training a classifier with the data collected while the robot interacts with the environment. The relevance map indicates the parts of the visual scene that are relevant for the robot with respect to an action. "Relevant parts" means parts of the  environment that have a high probability to produce an expected effect after having applied a given action, for instance moving this part of the environment when touching it through a push primitive.

The exploration is sequential and follows 5 main steps: 
\begin{itemize}
    \item The visual scene is over-segmented using Voxel Cloud Connectivity Segmentation \cite{Papon2013}. This method segments a 3D pointcloud into small regions of same size. Then, visual features are extracted from each segment.
    \item The \textit{relevance map} attributes to each segment a relevance weight computed using the prediction of the classifier trained online.
    \item Based again on the classifier, a \textit{choice distribution map} is computed which represents the probability of each segment to be chosen as the next interaction target.
    \item An action primitive is applied with the center of the chosen segment as target.
    \item Finally, an effect detector is applied to label the visual features of the selected segment. Detected effects are labelled to 1, otherwise a label equal to 0 is attributed.
\end{itemize}

By following these 5 steps, the robotic system builds a dataset of labeled samples on which a classifier is trained online. 

\begin{figure}[ht]
\centering
\subfloat[Simplest set-up used for the experiment. A toy workbench with 3 movable cars.]{
\includegraphics[width=.9\linewidth]{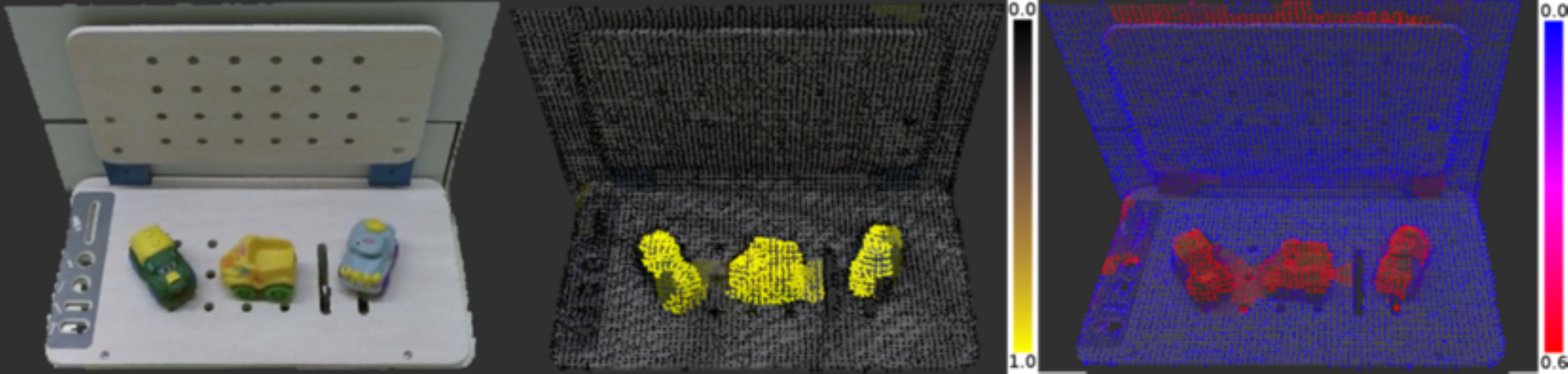}
}\\
\subfloat[A more complicated set-up. A toy workbench with an object fixed on the horizontal panel and 3 movable cars.]{
\includegraphics[width=.9\linewidth]{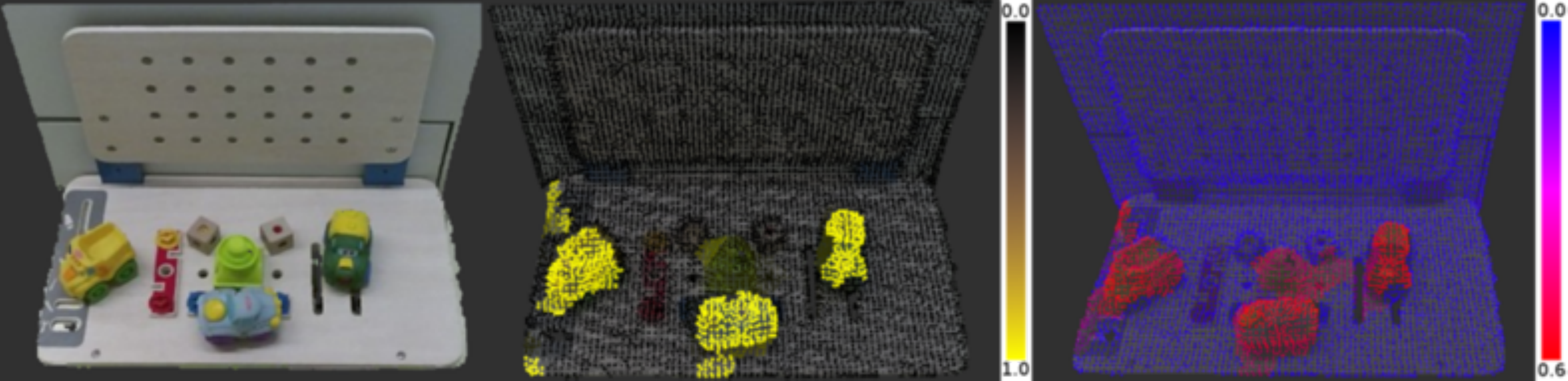}
} \\
\caption{Two examples of relevance map learned during an exploration with a push primitive. From left to right: a colored 3D pointcloud, a relevance map on the left point cloud and the accumulated choice distribution maps during the whole exploration. Adapted from \cite{goff2019bootstrapping}.}
\label{fig:rm_cdm}
\end{figure}

The proposed approach was tested on both a Baxter \cite{goff2017segmenting} and a PR2 \cite{goff2019bootstrapping} robots. The experiments were conducted on two set-ups with a push primitive. To detect if an effect occurred, a change detector compares the pointclouds before and after the interaction. If a targeted segment is part of the difference pointcloud, it means that something has moved. In this context, the relevance map represents the areas of the environment that the robot can move. Examples of obtained results are shown in Figures \ref{fig:rm_seq} and \ref{fig:rm_cdm}.

Figure \ref{fig:rm_seq} shows a sequence of relevance maps at different moments of the exploration. After the first interaction (i.e. only one sample in the dataset),  the map is uniform. From the $50^{th}$ interaction, the map begins to show a meaningful representation. An important feature of the classifier is its ability to give meaningful predictions with few samples. Therefore, the exploration is efficiently directed after a few collected samples. The exploration process is focused on complex areas, i.e. areas which carry a lot of information, as shown on left part of \figurename~\ref{fig:rm_cdm}. 

Several relevance maps relative to different action primitives can be learned. This approach  was tested  with a push primitive, a push-button primitive and a lift primitive \cite{goff2019building}. Each of these relevance maps is a representation of the environment depending on the action and on the possible effect considered during exploration. In other words, a relevance map implements an affordance \cite{gibson1979}. These relevance maps are finally merged into a new perceptual map, called \textit{affordance map}. An affordance map gives to the robot a rich perception of which action could be applied and where they could be applied. 

\begin{figure}[!ht]
\centering
\subfloat{
\includegraphics[width=.9\linewidth]{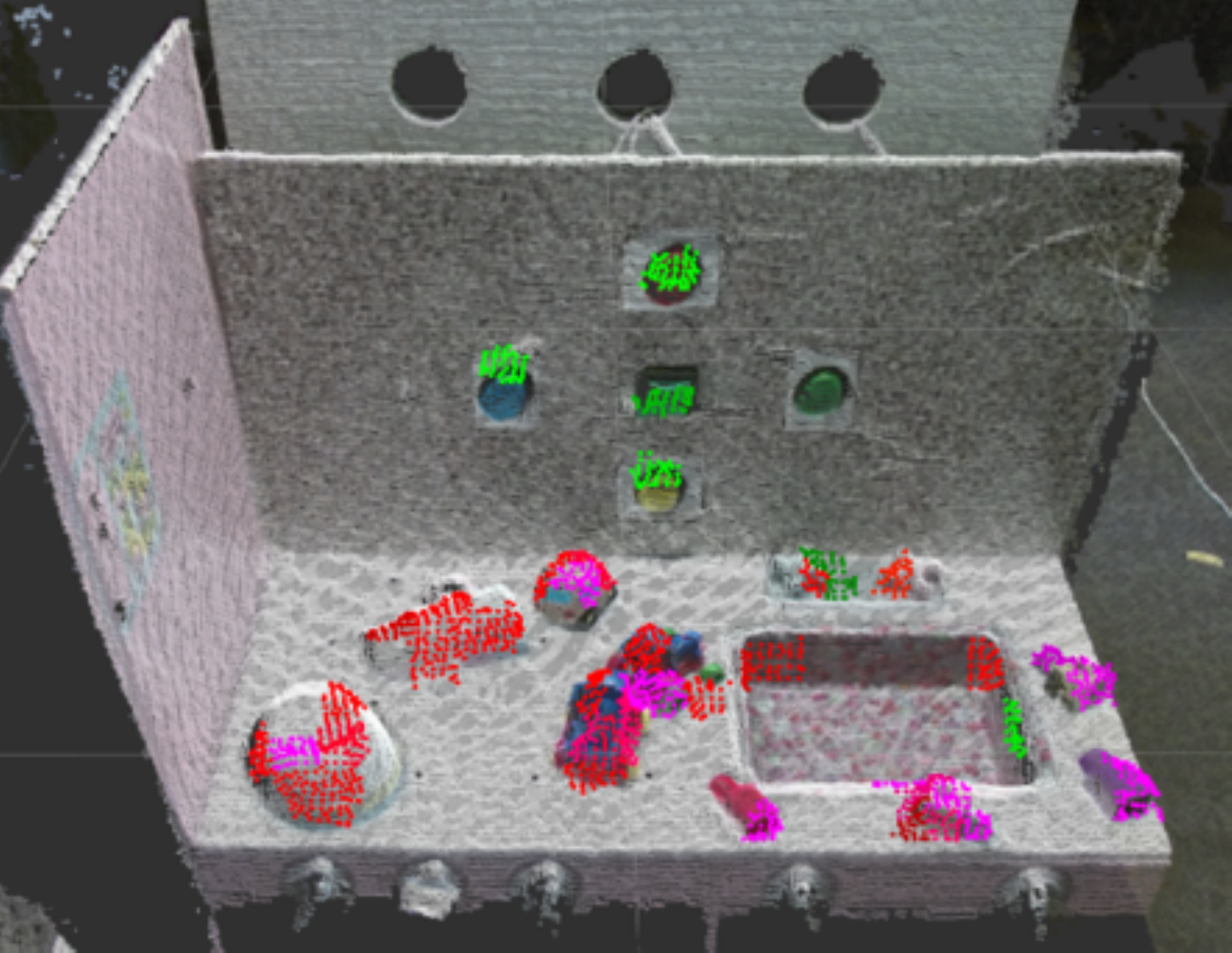}
}\\
\subfloat{
\includegraphics[width=.5\linewidth]{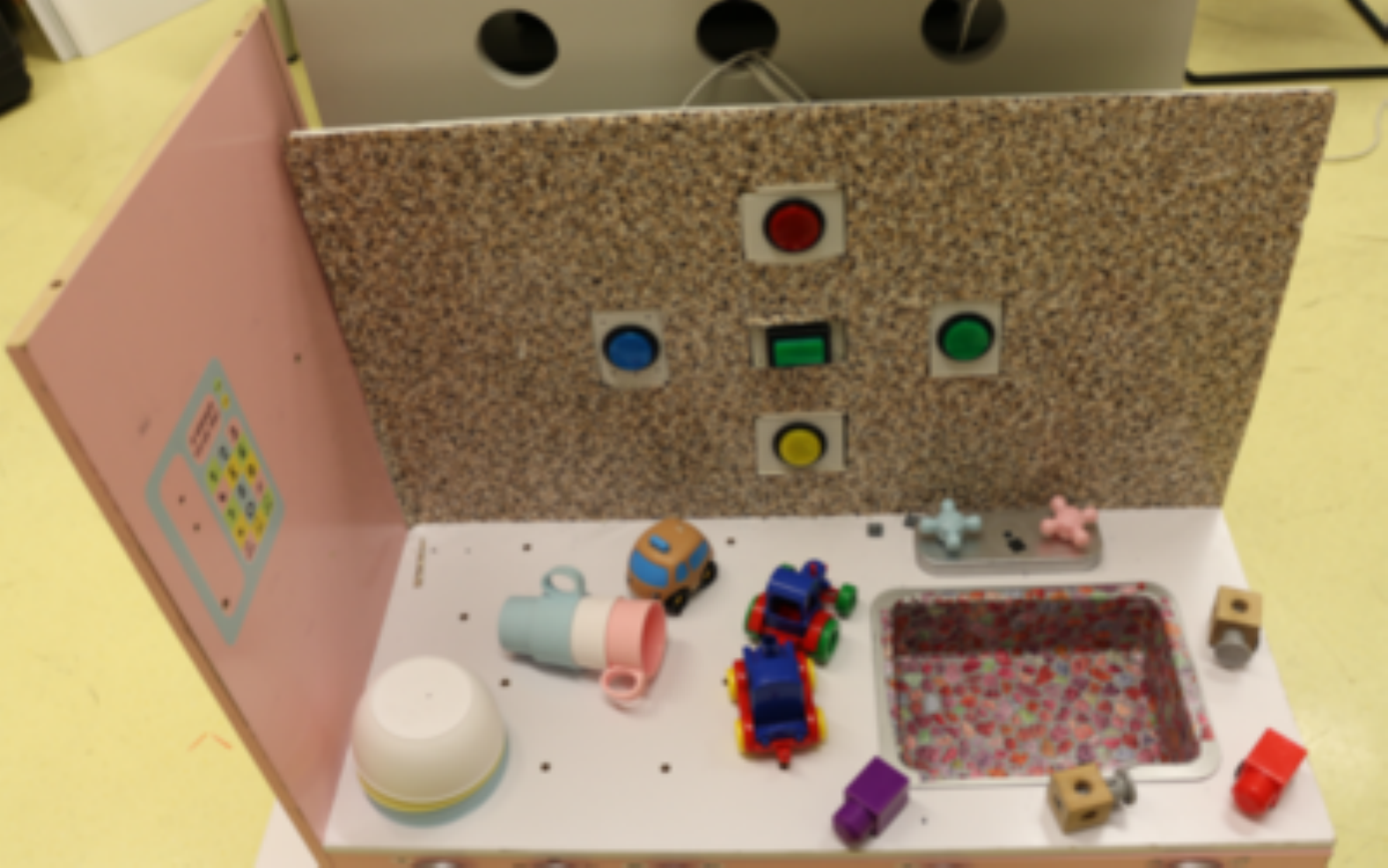}
}  
\caption{Affordance map of liftable objects (in purple), activable push-buttons (in green) and pushable objects affordances (in red). Only areas classified with a probability of afforded action above 0.5 are represented in the figure. The bottom picture represents the environment from which the affordance map has been extracted. Adapted from \cite{goff2019building}.} 
\label{fig:aff_map}
\end{figure}

An example of affordance map is shown in \figurename~\ref{fig:aff_map}. The push-buttons identified in green by our system do not overlap with the pushable and liftable objects identified in red and purple. Thus, the classifier is able to learn different concepts. Also, only small objects are identified as liftable and pushable, and the biggest objects are identified as only pushable. 

\paragraph{Object oriented babbling}

The second step consists in building object models on the basis of the segmented maps acquired during the babbling phases. The description of this module is out of the scope of this article. The module can, for instance, rely on the method proposed by \cite{biegelbauer2010model}. 
It aims at providing a 3D model of the environment for learning processes described in the next section.

\subsection{Learning to manipulate objects}

\begin{figure}[!ht]
\centering
\includegraphics[width=0.8\linewidth]{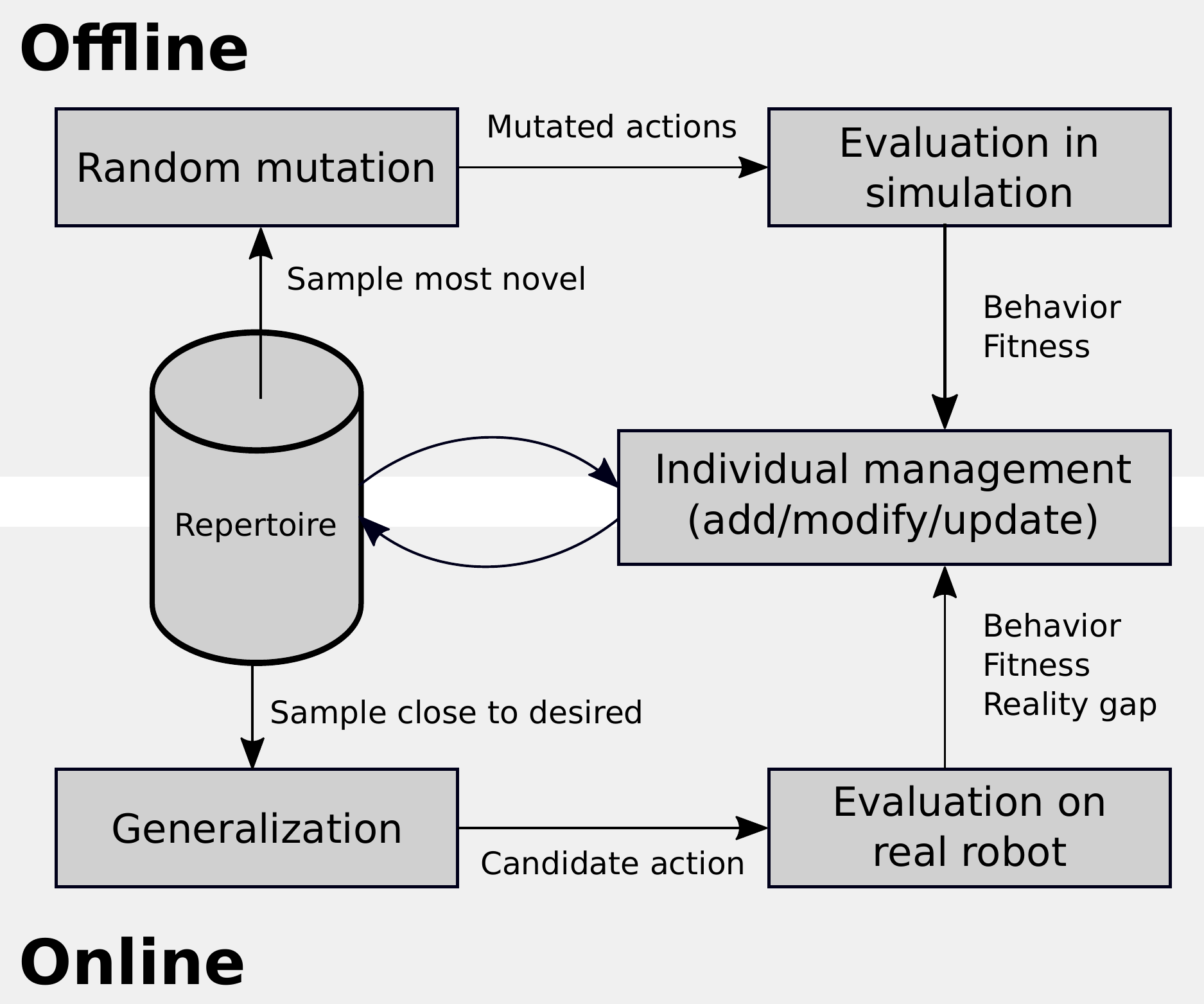}
\caption{Overview of the object-oriented skills learning module. A quality-diversity algorithm (in blue) is run offline (using a simulation) to build a repertoire of motor skills able to produce diverse behaviors. When the robot must perform a given behavior (in red), it then samples the archive for the skills producing the closest behaviors, and uses them to build a candidate skill expected to reach the target point in the outcome space. If this skill fails due to the reality gap, the error is evaluated and used to adapt the candidate skill, and to update neighboring skills. Adapted from \cite{kim2019exploration}.}
\label{fig:qd_overview}
\end{figure}


With the previously described bootstrapping modules, the system can acquire a model of its environment, the various objects it contains and their properties. In order to solve tasks involving those objects, the robotic system must now learn motor skills to manipulate them. This raises the challenge of exploring and mapping the action space of the robot to build motor skills able to engage them. In an open-ended learning context, the objects and environment can vary, and it is therefore of great importance that the same methods can handle different setups and use no prior knowledge other than that provided by the bootstrap module. Furthermore, instead of reaching a specific goal for which finding a single policy would be enough, as is typical in a reinforcment learning paradigm, we want the system to be able to tackle different tasks requiring various skills. 

A way to address this issue is to define not a goal but a space representing the controllable state of the environment, often called behavior space~\cite{lehman_abandoning_2010}, goal space~\cite{Pere2018} or outcome space~\cite{sigaud2019policy}, and to learn a wide repertoire of motor skills able to reach many points in this space. As those exploration algorithms tend to be sample-inefficient, they are usually used in simulation, which is possible in the present context considering the knowledge acquired from the bootstrap phase, but introduces a further challenge in the form of the reality gap~\cite{Jakobi1995}, where policies learnt in simulation must be transferred to the real robot. Our approach~\cite{kim2019exploration} uses a Quality-Diversity (QD) algorithm~\cite{pugh2016quality,cully2017quality} to build such a skill repertoire, and a generalization approach based on a local linear model of the mapping from the action parameter space to the outcome space to adapt those skills to real robot control. This process is summarized in \figurename~\ref{fig:qd_overview} and detailed below. We evaluate this approach for two different problems (throwing a ball at various targets and manipulating a joystick) and show that in both cases, it is able to learn repertoires of diverse skills, to address the reality gap issue, and to generalize to new policies, resulting in efficient control of the outcome space.

\subsubsection{Offline learning of skill repertoires}

\begin{figure}[ht]
\centering

\subfloat[Example of diverse throwing trajectories]{\label{fig:diverse_throw}
\includegraphics[width=.99\linewidth]{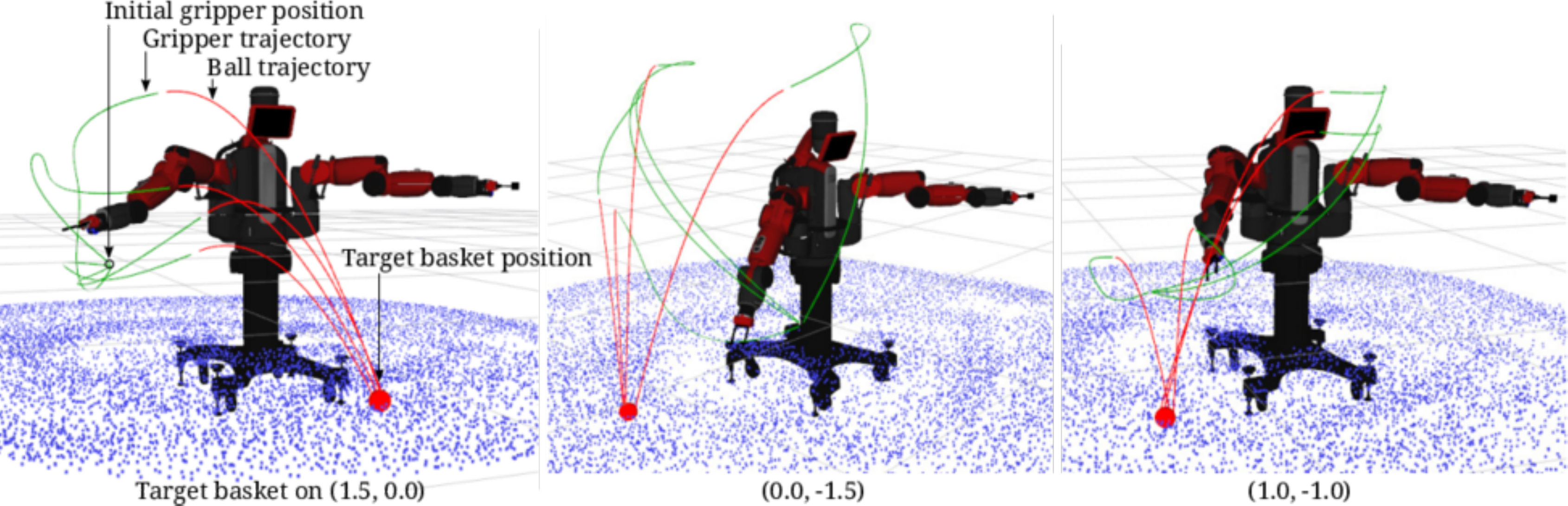}}\\

\subfloat[Random baseline]{\label{fig:throw_random_archive}
\includegraphics[width=.49\linewidth]{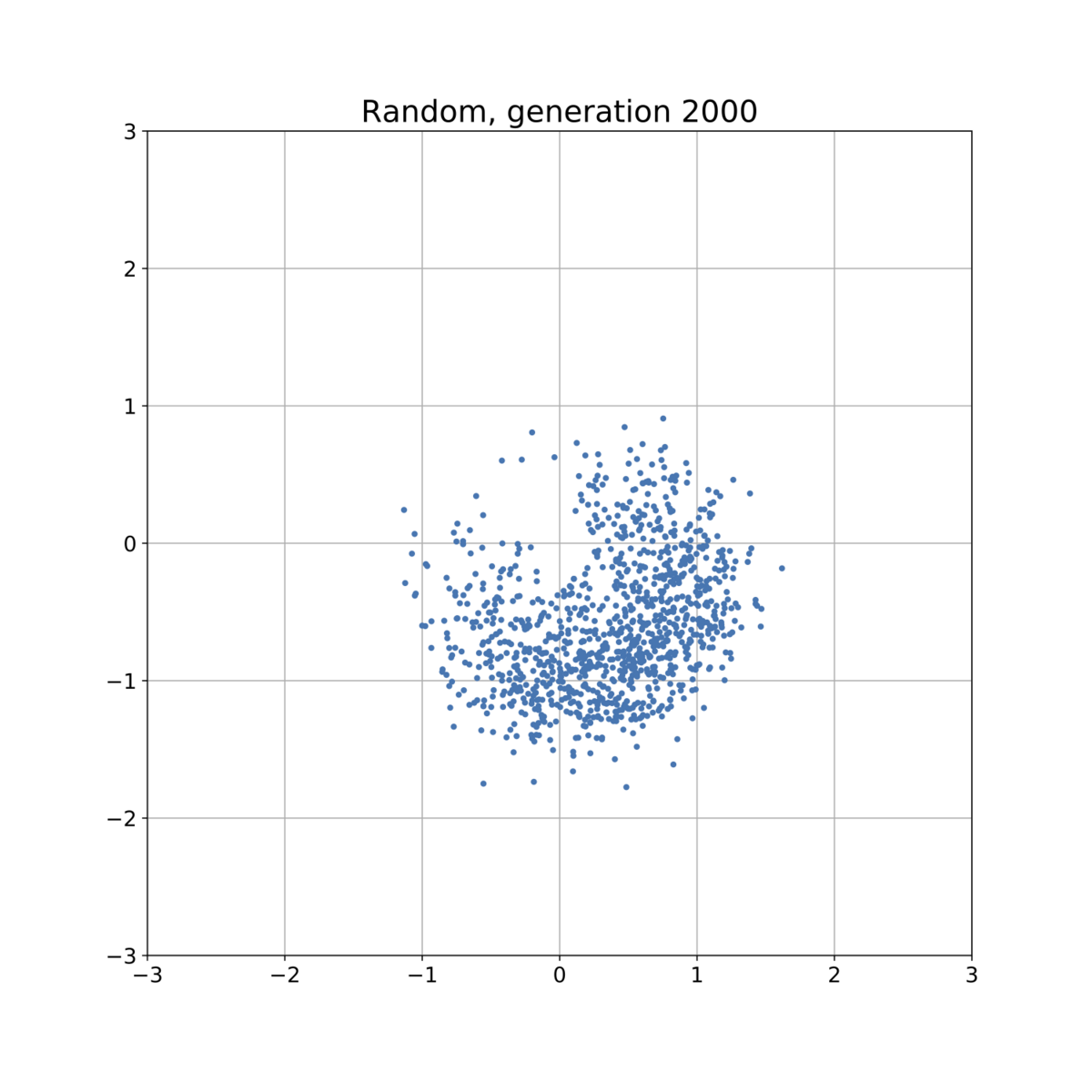}
} 
\subfloat[QD search]{\label{fig:throw_qd_archive}
\includegraphics[width=.49\linewidth]{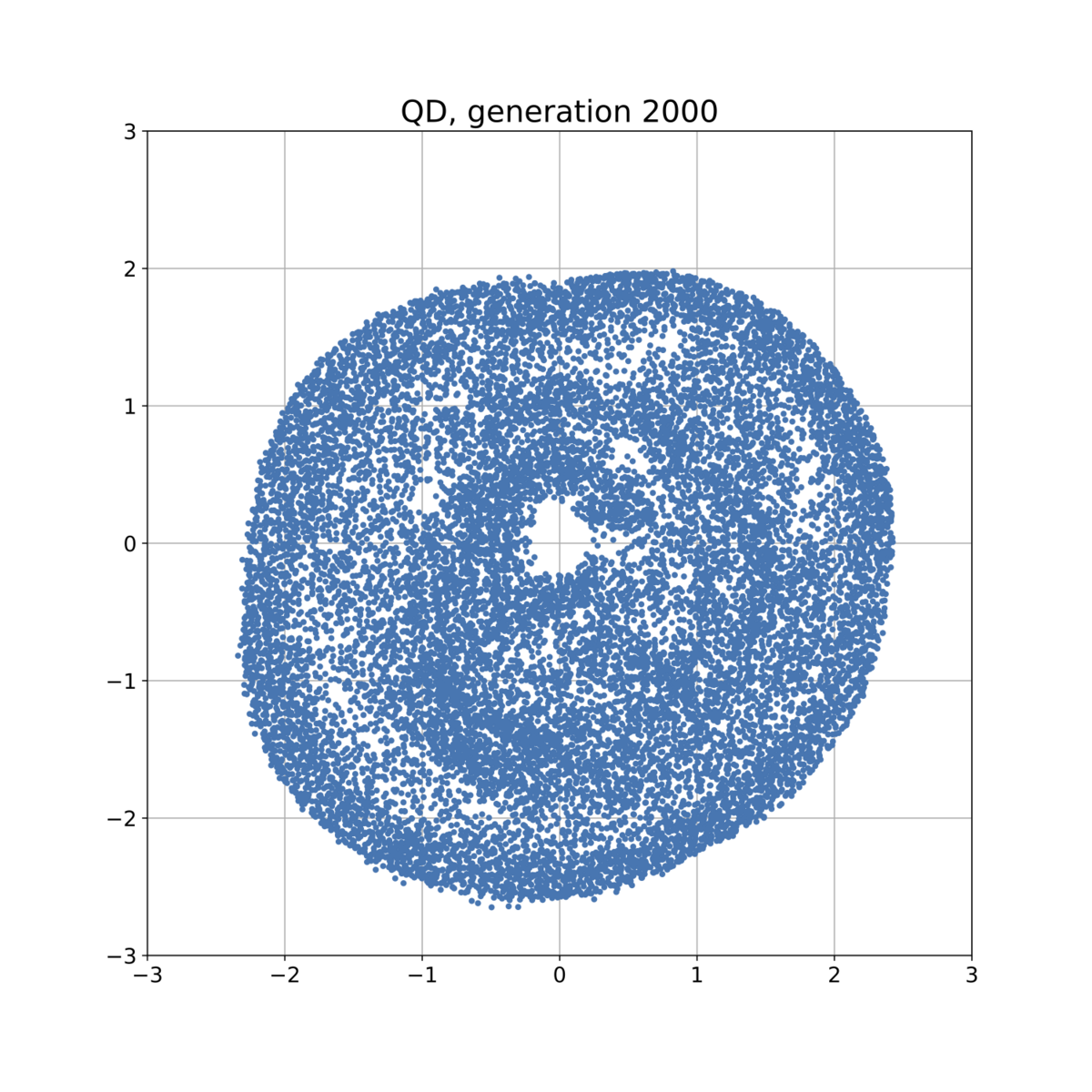}
} 

\caption{Skill repertoires built by the random baseline (\ref{fig:throw_random_archive}; \num{1085 \pm 26} points) and the QD search (\ref{fig:throw_qd_archive}; \num{14473 \pm 1619} points) for the ball throwing problem. Each blue point is the contact point of the ball with the ground. QD search was run for 2000 generations. For the random baseline an equal number of actions were uniformly sampled in the action space. Adapted from \cite{kim2019exploration}.}
\label{fig:qd_throw}
\end{figure}

\begin{figure}[ht]
\centering

\subfloat[Example of diverse joystick manipulation skills. Note that similar joystick positions can be reached by different movements.]{\label{fig:diverse_joy}
\includegraphics[width=.99\linewidth]{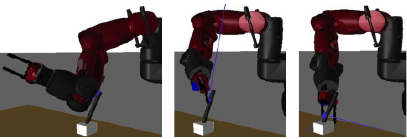}} \\

\subfloat[Random baseline]{\label{fig:joy_random_archive}
\includegraphics[width=.49\linewidth]{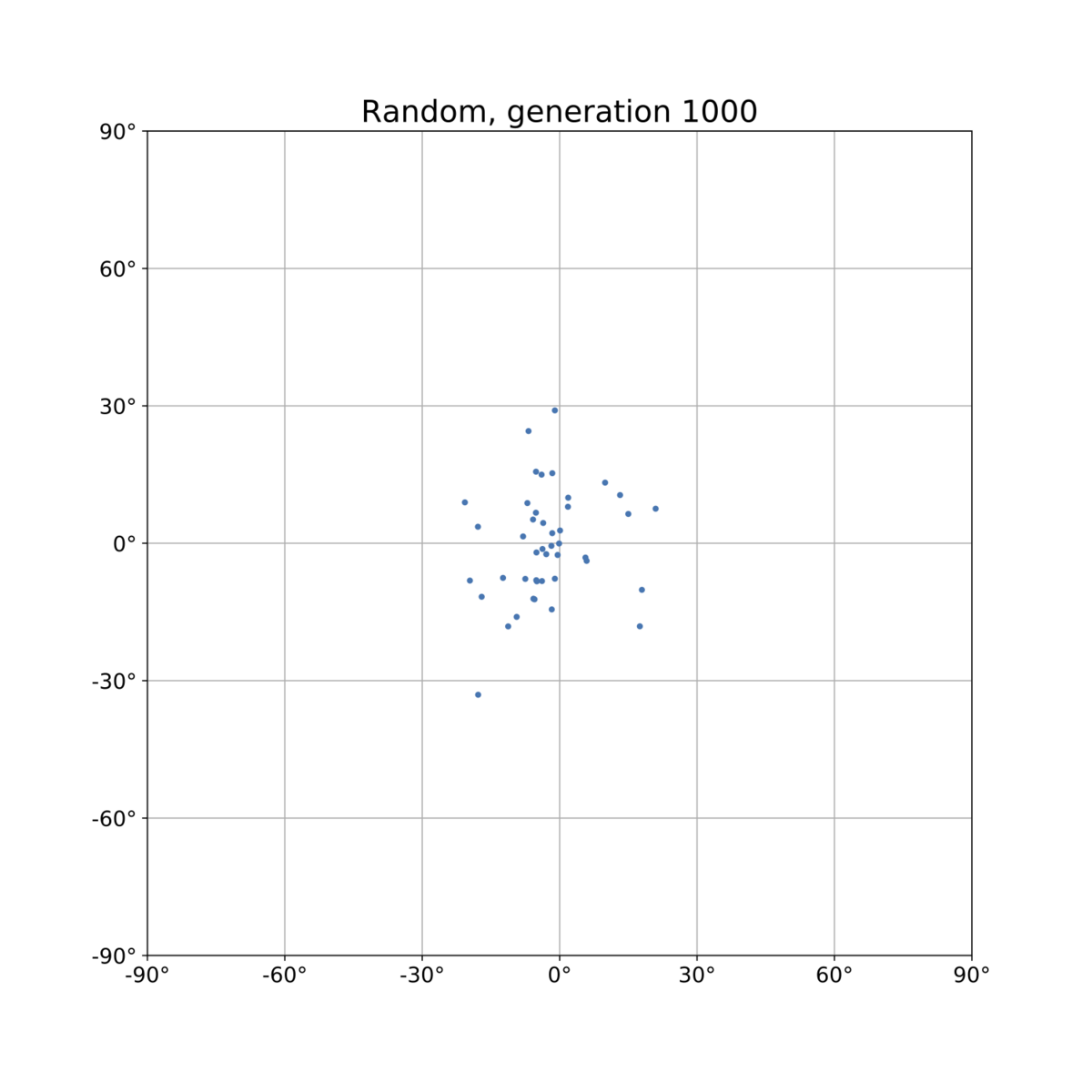}
} 
\subfloat[QD search]{\label{fig:joy_qd_archive}
\includegraphics[width=.49\linewidth]{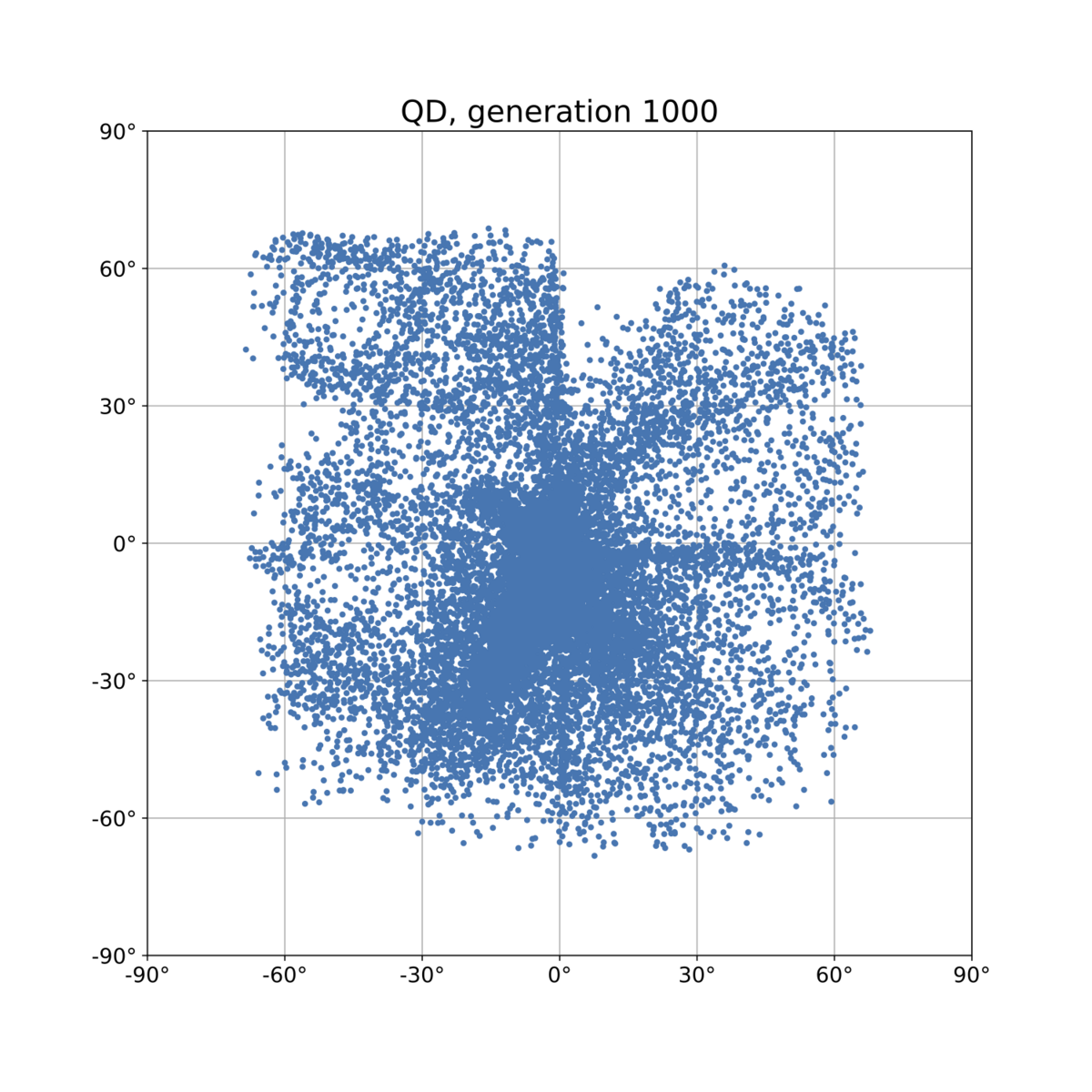}
} 

\caption{Skill repertoires built by the random baseline (\ref{fig:throw_random_archive}; \num{34 \pm 3} points) and the QD search (\ref{fig:throw_qd_archive}; \num{15532 \pm 3329} points). Each blue point is a final joystick position QD search was run for 1000 generations, for the random baseline an equal number of actions were uniformly sampled in the action space. Adapted from \cite{kim2019exploration}.}
\label{fig:qd_joy}
\end{figure}

For both problems, the system was tested on a Baxter robot controlled by simple parameterized motion primitives based on a third order polynomial, whose parameters constituted the action space (see~\cite{kim2019exploration} for details). For the ball throwing problem, a ball was initially placed in the robot's gripper and the studied outcome space was the 2D position of the ball when it reached the ground plane (\figurename~\ref{fig:diverse_throw}). For the joystick manipulation problem, a joystick was placed on a table in front of the robot and the outcome space was its final pitch and roll (\figurename~\ref{fig:diverse_joy}).
The quality metric used for ball throwing was torque minimization, and skill robustness to small perturbations for joystick manipulation. Skills were added to the repertoire if they had no close neighbor, or replaced their closest neighbor if they had a higher quality score.
Evaluation was done using the DART simulator. Results show that the quality diversity algorithm is able to learn a skill repertoire that densely covers the reachable outcome space for ball throwing (\figurename~\ref{fig:throw_qd_archive}), and another skill repertoire to reach a large and diverse set of final positions for joystick manipulation (Fig.~\ref{fig:joy_qd_archive}) whereas a random baseline (\figurename~\ref{fig:throw_random_archive},~\ref{fig:joy_random_archive}) results in much more limited exploration of the outcome space\footnote{Only one run of the method out of 26 for ball throwing and 12 for joystick manipulation is shown in Figs.~\ref{fig:qd_throw} and \ref{fig:qd_joy}; see~\cite{kim2019exploration} for full results and analysis.}.

\subsubsection{Online generalization and adaptation by local linear Jacobian approximation}

Using a skill repertoire generated by the QD algorithm to control the robot in the real world raises two challenges: first, the skills may have different outcomes in reality than in the simulated environment (the reality gap problem); second, despite densely covering the outcome space, the repertoire is still finite, and may not contain the skills to reach some specific points in that space. Our local linear Jacobian approximation method, similar to that of \cite{Baranes2013goalexploration}, tackles both issues. It proceeds as follows (with $\mathcal{A} = \{(\bm{\theta}_i,\bm{b}_i)\}_{i=1,\ldots,N}$ the repertoire containing $N$ action parameters $\bm{\theta}_i \in \mathcal{G}$ and their outcomes $\bm{b} \in \mathcal{B}$): 

\begin{itemize}
\item For an arbitrary target point $\bm{b}^* \in \mathcal{B}$, find the closest point $\bm{b}_c$ in the repertoire and its corresponding $\bm{\theta}_c$.
\item Find the $K$ nearest neighbors to $\bm{\theta}_c$ in the repertoire, and their corresponding outcomes.
\item Use those $K$ $(\bm{\theta}_i,\bm{b}_i)$ samples to estimate $\bm{J}_{\bm{\theta_c}}$ the Jacobian matrix\footnote{the matrix $\bm{J}_{\bm{\theta}}$ such as $\Delta\bm{b} = \bm{J}_{\bm{\theta}} \Delta \bm{\theta}$} at $\bm{\theta}_c$ by the least squares method.
\end{itemize}

This estimation $\widetilde{\bm{J}}_{\bm{\theta_c}}$ can be used to define a local linear model between the action parameter space to the outcome space -- and a matching local inverse model, by pseudo-inverting the matrix. Although the global mapping of the action parameter space to the outcome space is highly non-linear for the considered problem, it is smooth at most points and the skill repertoire is dense enough to define a good linear approximation in most regions. The local linear model can then be used to solve the aforementioned issues:

\begin{itemize}
\item \textbf{Generalization:} using the local inverse model, compute a candidate action $\widetilde{\bm{\theta}}^*$ which is expected to reach $\bm{b}^*$, and try it on the robot;
\item \textbf{Reality gap crossing:} if the candidate action does not reach $\bm{b}^*$ accurately enough, record the point $\widetilde{\bm{b}}^*$ reached and compute the error $\Delta \bm{b}^* = \widetilde{\bm{b}}^* - \bm{b}^*$. The pseudo-inverted Jacobian estimation can then be directly used to compute a correction $\Delta \bm{\theta}^*$ to the action to apply to $\widetilde{\bm{\theta}}^*$ to reduce the error. This process can be iterated if needed, until the reality gap has been crossed.
\end{itemize}

\begin{figure}[!ht]
\centering
\subfloat{\label{fig:real_throw}
\includegraphics[width=.45\linewidth]{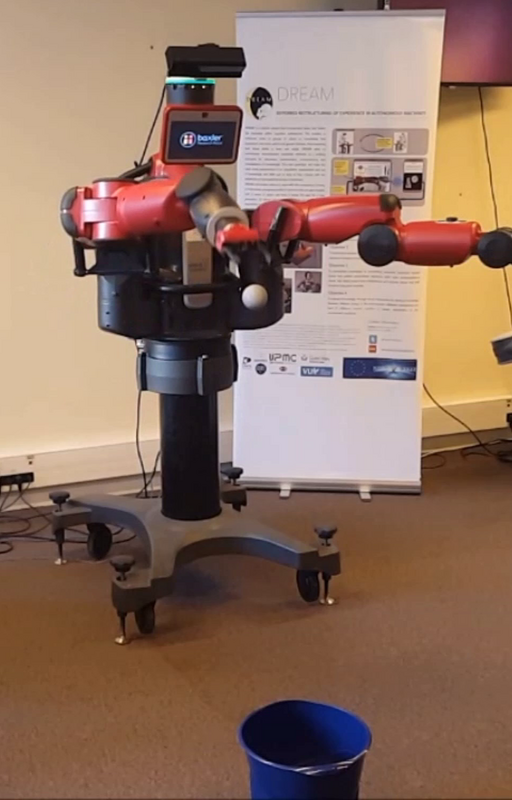}
}
\subfloat{\label{fig:real_joy}
\includegraphics[width=.45\linewidth]{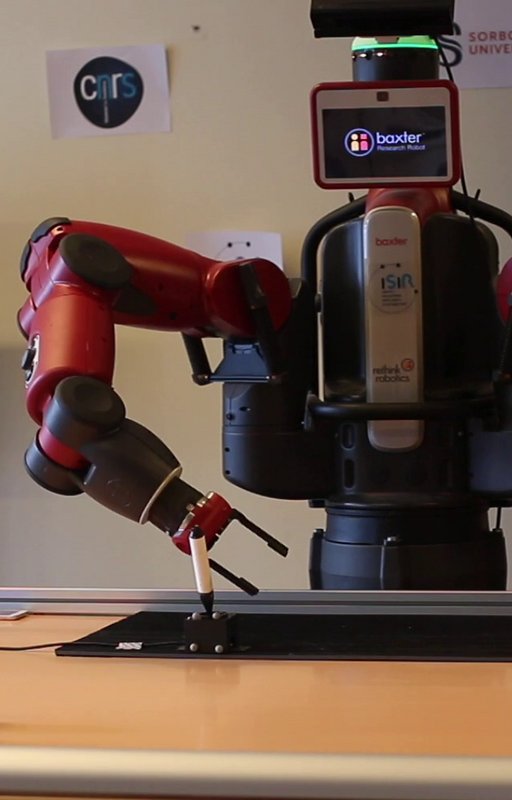}
}  
\caption{The real robot performing ball throwing towards a basket (left) and joystick manipulation (right) based on the action repertoires built in simulation.}
\label{fig:real_robot_repertoire}
\end{figure}

Reality gap crossing was quantitatively evaluated in simulation, with a large simulated reality gap. In both conditions, most actions initially failed due to the reality gap, but could be adaptated by the method in at most 4 iterations of the method in 89\% of cases for ball throwing, and 31\% of cases for the more difficult joystick manipulation task. Reality gap crossing was also tested on real robot (Fig.~\ref{fig:real_robot_repertoire}): over 50 trials on random target positions, only 8 required adaptation, and all 8 succeeded after a single iteration of the method \cite{kim2019exploration}.

\section{Building new action representations}
\label{sec:action_repr}

We would like the behavioural capabilities of our robot to be robust to  environmental perturbations. Unexpected changes in the environment may require using different actions to achieve the same effect, for instance to reach and grasp an object in changing clutter. A possible approach is to adapt the control policy through, e.g., obstacle localization and explicit re-planning \cite{stulp2009planning}. This approach requires to build a dedicated algorithm in which obstacle representation is given beforehand. This may raise an issue for open-ended learning. A more general purpose and open-ended alternative is to exploit a behavioural repertoire \cite{duarte2018repertoire} and extract from it an adequate policy \cite{cully2015robots}.

In the bootstrapping section presented earlier, we described  learning a repertoire using QD search, exemplified by the Baxter robot throwing a ball. In this case the repertoire spans the space of potential throwing targets. After this bootstrapping phase, one (or a small number of) throwing policies are stored for each potential target. To increase behaviour robustness through diversity, we need multiple diverse throwing movements \emph{for each potential target}. Thus, if a new obstacle appears, diverse behaviours can be tried until one succeeds. A behavior repertoire contains a finite and limited number of policies that can thus adapt only to a certain extent to new situations. 

\begin{figure}[htb!]
\begin{center}
    \includegraphics[width=\linewidth]{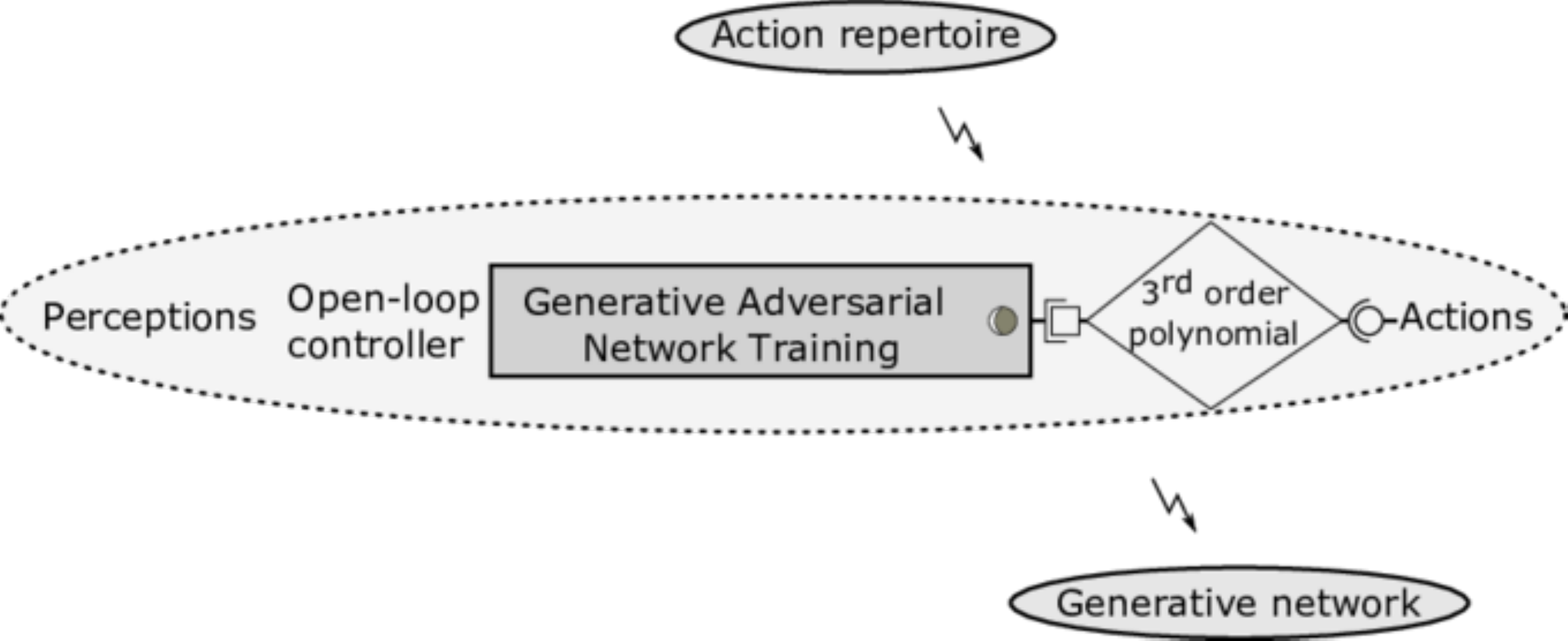}
\end{center}
\caption{\label{fig:gan_step}The proposed action redescription phase that builds a policy parameter generative network out of an action repertoire.}
\end{figure}

To go beyond this limitation, we present an action redescription process transforming the library-based representation obtained from the bootstrap phase into a new representation that is both more compact and more diverse --  through learning a generative adversarial network (GAN) \cite{goodfellow2014genAdversarialNets} over policies (\figurename~\ref{fig:gan_step}). GANs are neural networks suitable for learning generative models over complex high-dimensional data, typically images. In this case, we train a conditional GAN that accepts a movement (throwing) target as a condition, and generates diverse throwing movements that hit this target. Now, only the model parameters rather than controller library needs to be stored, and the available diversity is not limited to a fixed length controller library. By sampling the generative model over controllers, an unlimited number of distinct controllers can be obtained. Given a powerful generative model, these need not be simple perturbations of known controllers, but can encode novel solutions to the problem by drawing on diverse aspects of multiple training policies. 

\begin{figure}[ht!]
 \centering 
 \includegraphics[width=1.0\columnwidth]{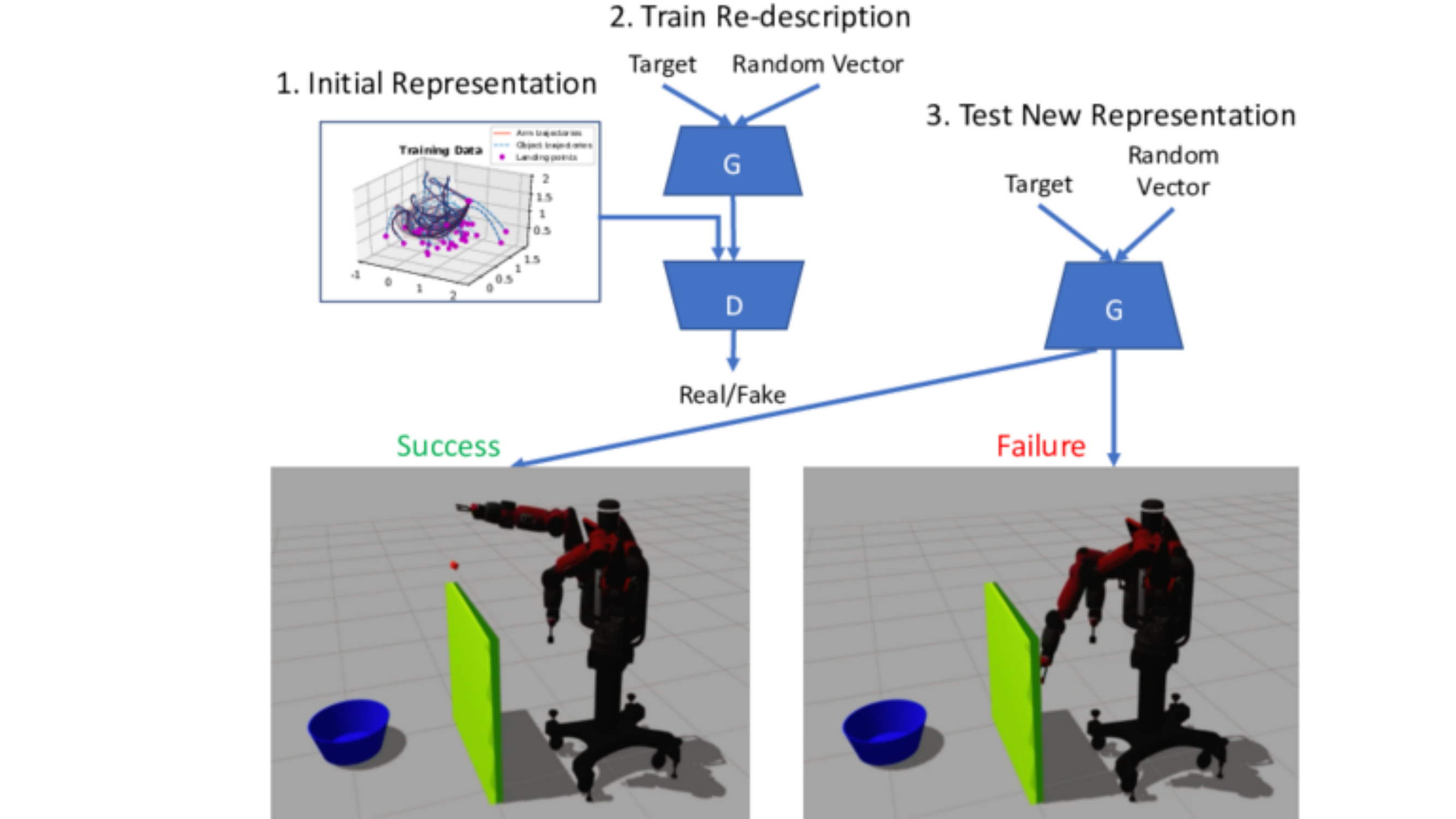}
 \caption{Action redescription for robustness through diversity. 1. The initial representation is the population of controllers from QD search. 2. GAN training produces a generative model over controllers. 3. For robust behaviours, the generative policy network is sampled until one is found that avoids the obstacle.}
 \label{fig:UE_Gan}
\end{figure}

Figure~\ref{fig:UE_Gan} illustrates our framework, with full details available in \cite{jegorova2018generative}. The policy representation here is a 15D vector of parameters defining a low-level open-loop velocity controller for throwing. We start with a set of controllers obtained from QD search. The diversity of this set mainly spans different throwing targets. We then train a target-conditional generative model for controllers by playing a min-max game with a generator and discriminator network. Once the generator network is trained, it maps a target coordinate on the floor, and a random vector to a new controller. Sampling this random vector for a fixed target vector generates diverse ways of throwing to the same target. In the case of an obstacle, controllers can be sampled until one is found where neither the ball nor the arm collide, and the ball hits the target. This is illustrated in \figurename~\ref{fig:UE_Gan} where two throwing samples are drawn, and the underhand throw fails while the overhead one succeeds. 

\begin{figure}[ht!]
 \centering 
 \includegraphics[width=0.8\columnwidth]{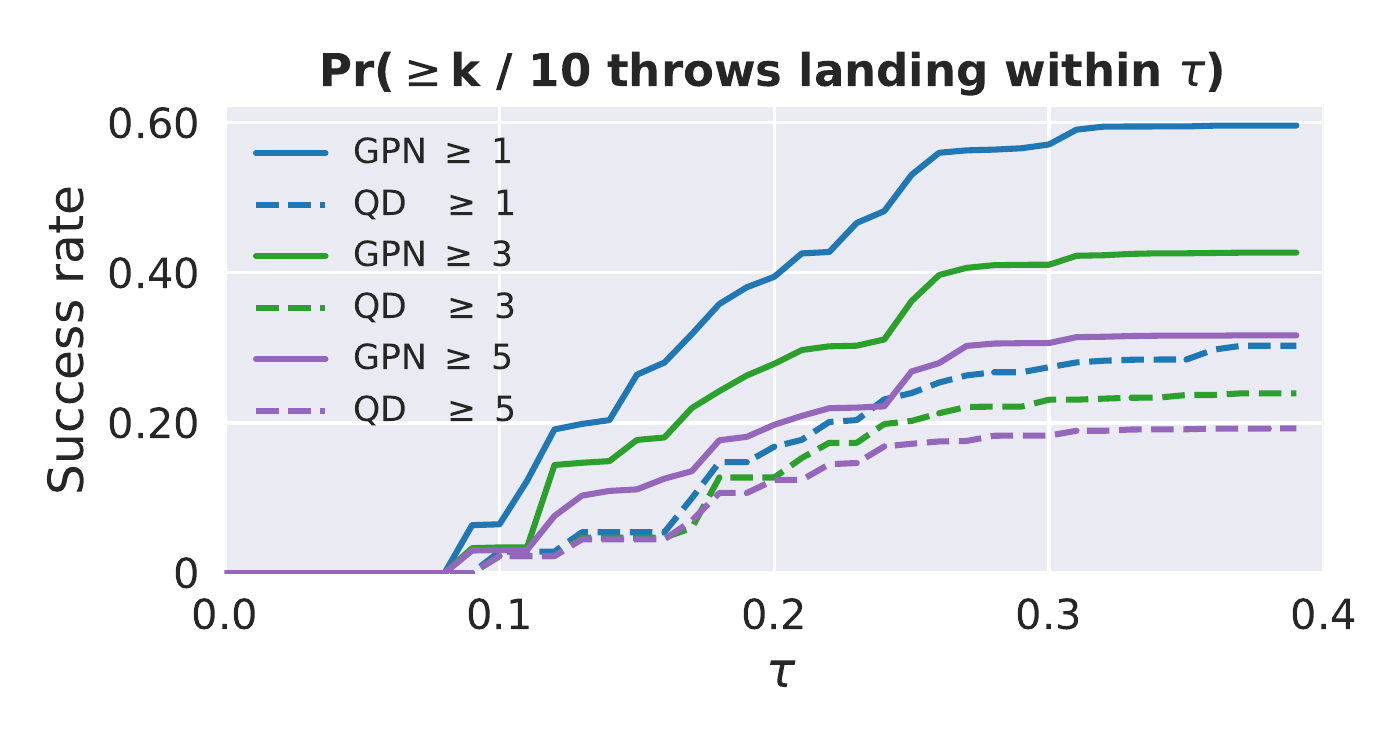}
 \caption{Quantitative evaluation of throwing in the presence of obstacles. Our redescription improves in terms of probability of achieving $k$ hits out of 10 for various target radii $\tau$. Adapted from \cite{jegorova2018generative}.}
 \label{fig:UE_Gan2}
\end{figure}

For quantitative evaluation, we perform the throwing task averaging over a large number of target positions, randomly placed obstacles, and multiple diverse throwing attempts in each target-obstacle configuration. Our goal is that out of $N$ throwing attempts in each configuration, at least $k$ of them should hit the target. The results in \figurename~\ref{fig:UE_Gan2} show probability of $k$ hits out of $N=10$, as a function of how close to the target a ball should land to be considered a hit. We can see that, as expected, the success rate depends on the stringency of the hit criterion. More interestingly, the proposed redescription increases this rate (solid vs dashed lines), at several values of $k$. Thus, this action redescription succeeds in increasing robustness via increased diversity, while also compressing the prior bootstrap representation.

\section{Transferring knowledge}

\subsection{From one task to another}

\begin{figure}[htb!]
\begin{center}
    \includegraphics[width=0.8\linewidth]{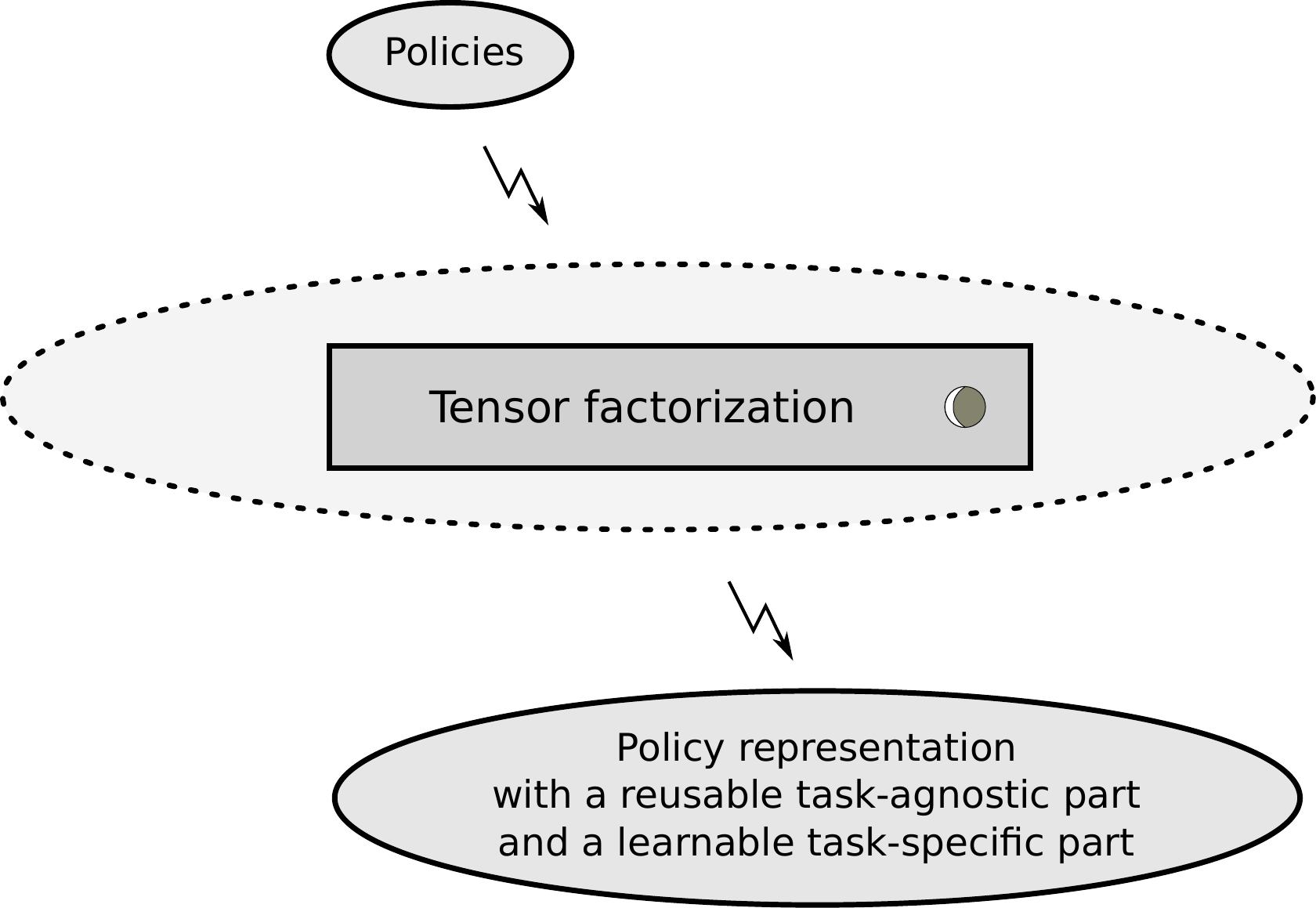}
\end{center}
\caption{\label{fig:tensor_factorization}Tensor based knowledge transfer relies on a tensor factorization to perform a policy redescription. It builds, from a set of policies, a new policy representation that includes a reusable part and a learnable part. The use of a task-agnostic, reusable knowledge allows further learning to be faster.}
\end{figure}

Learning an individual robot control task from scratch usually requires a large amount of experience and may physically damage the robot. This has motivated a fruitful line of research into transfer learning, which aims to bootstrap the acquisition of novel skills with knowledge transferred from previous acquired skills \cite{taylor2009transfer}. For open-ended learning, we would like to transfer knowledge from a lifetime of previous tasks rather than a single source task. One potential  way to realize this is to model all tasks as laying on a low dimensional manifold \cite{agarwal2010learning}. Based on this assumption, \cite{kumar2012learning} construct a transferrable knowledge base (the manifold) from linear policies through matrix decomposition methods, in the case of supervised learning tasks. \cite{yang2017deep} extend this approach to non-linear policies, represented by deep neural networks, by stacking policies into 3-way tensors modeled by their low-rank factors under a Tucker decomposition assumption. This manifold-based transfer approach has been shown successful in learning simple linear robot control tasks, such as cart-pole \cite{ruvolo2013ella}. Here we further look into more complicated tasks and propose a knowledge transfer approach for learning novel non-linear control tasks.

\begin{figure}
\centering
\includegraphics[width = 0.95\columnwidth]{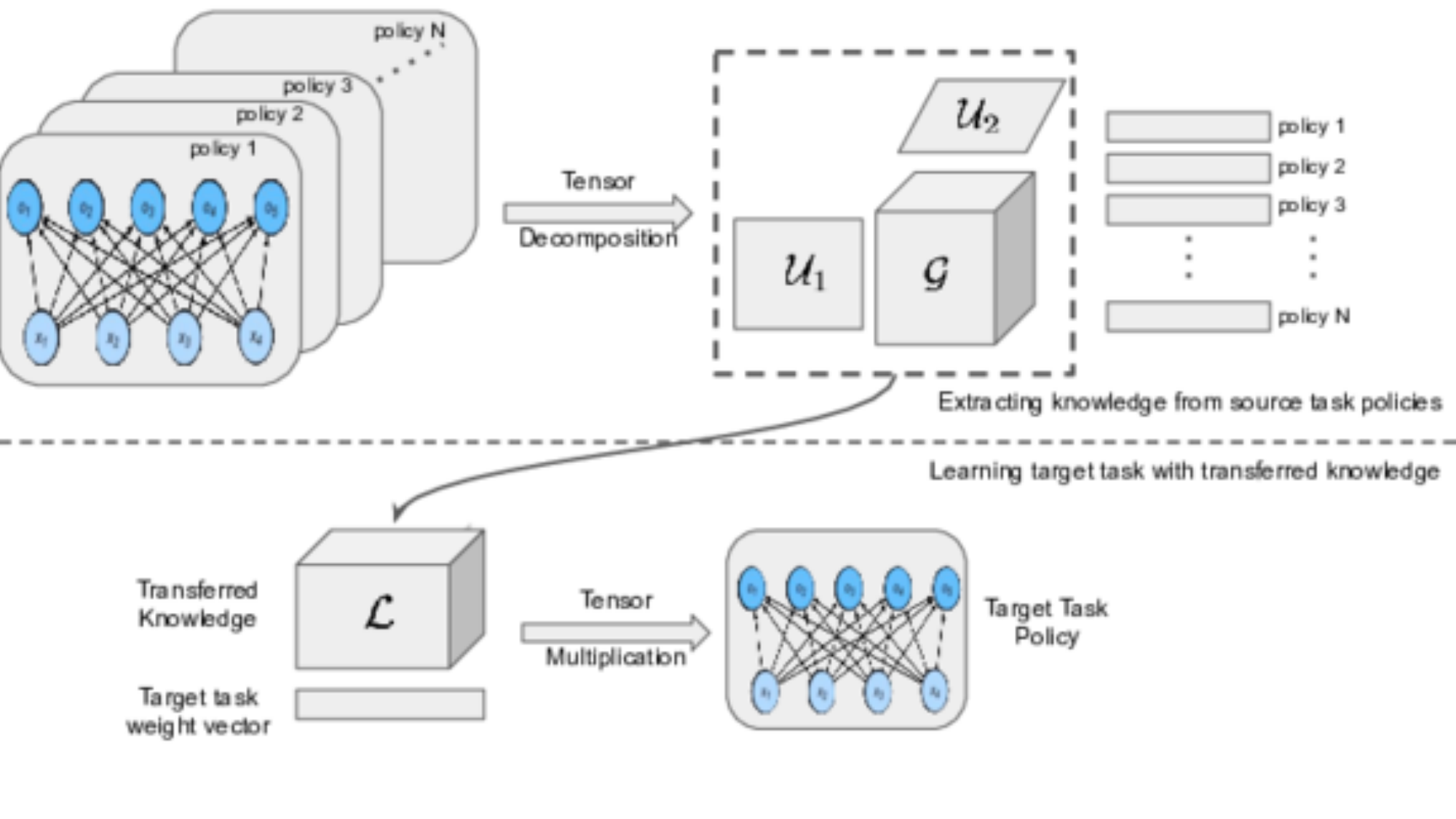}
\caption{A schematic diagram of tensor based knowledge transfer. Source task policies are stacked and decomposed into a task agnostic knowledge base tensor $\mathcal{L}$ and one task-specific weight vector for each task. Given a target task, the agent learns a target task weight vector and fine-tunes the knowledge base to reconstruct a target task policy.}
\label{fig:transfer_flow}
\end{figure}


\begin{figure}
\centering
\subfloat[Pusher]{\includegraphics[width=0.33\columnwidth]{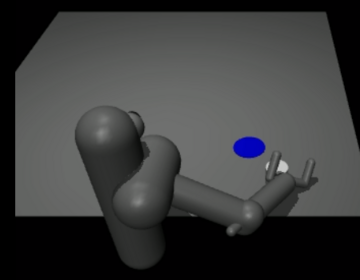}}
\subfloat[Thrower]{\includegraphics[width=0.33\columnwidth]{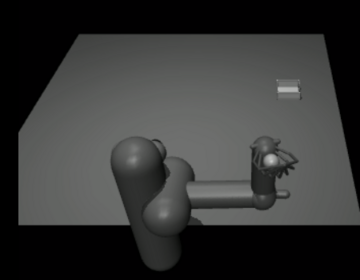}}
\subfloat[Striker]{\includegraphics[width=0.33\columnwidth]{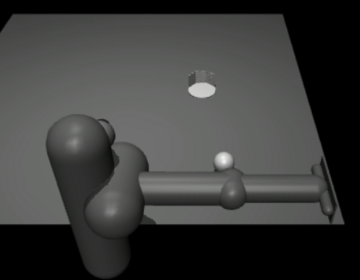}}
\caption{Illustrative figures of three robot manipulation tasks used to evaluate tensor-based cross-task knowledge transfer}.
\label{fig:mujoco_tasks}  
\end{figure}

We consider learning a policy $\pi_{n+1}$ for task $n+1$ given the policies of the previous $n$ learned tasks (\figurename~\ref{fig:transfer_0}) on the basis of an extracted, task-agnostic and thus reusable part and a learnable, task-specific part (\figurename~\ref{fig:tensor_factorization}). We use generic multi-layer perceptron (MLP) networks to model the policy for each task that maps instantaneous proprioceptive state to control torques at each joint. For each network layer, we stack the policies from source tasks into a 3-D tensor. To abstract the previous knowledge, we then factorise the tensor into a task-agnostic knowledge base $\mathcal{L}$ and task-specific task weight vectors with Tucker decomposition. For training the novel $n+1^{th}$ task, the agent alternates between learning the task-specific parameter vector and fine-tuning the task-agnostic tensor. The transfer procedure is illustrated in \figurename~\ref{fig:transfer_flow}, with full details available in \cite{zhao2017tensor}.

\begin{figure}[t]
    \centering
    \includegraphics[width=0.7\columnwidth]{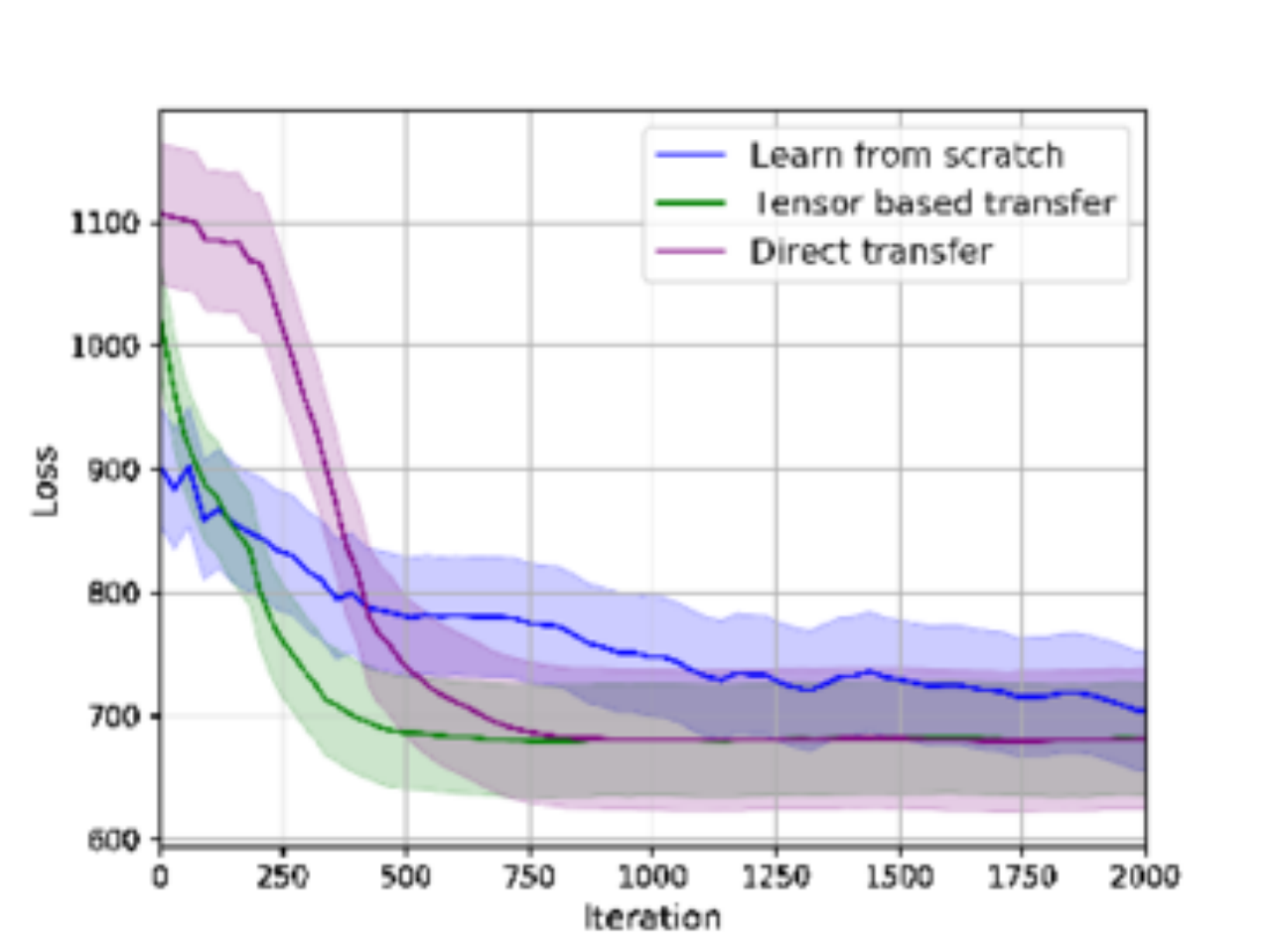}
    \caption{Learning curve for target task (Striker) with knowledge transferred from the source task policies (Thrower and Pusher).}
    \label{fig:striker_learning_curve}
\end{figure}

We evaluate our method with three simulated robot manipulation tasks as illustrated in \figurename~\ref{fig:mujoco_tasks}: Mujoco's Pusher, Thrower and Striker. We take pusher and thrower as source tasks, and striker as target task. We then evaluate the performance of reinforcement learning of the striker task with CMA-ES, comparing three alternatives for task-transfer: (i) learning from scratch without transfer, (2) directly transferring a randomly sampled source policy (pusher, or thrower) and fine-tuning for the target-task, (3) Our tensor-based transfer method. As shown in \figurename~\ref{fig:striker_learning_curve}, our method learns faster compared to both baselines. This is due to the ability to leverage the transferred abstract task agnostic knowledge obtained by re-representing the source policies through tensor factorization, which in this case corresponds to smooth movement primitives.

\subsection{From short term memory to long term memory}

In the work presented so far, the focus was on state or action representation learning, with the modules necessary to bootstrap it. Once this knowledge has been acquired, the robot has several different MDPs at its disposal, each, with its own state space, action space, reward function and policies. A fundamental question is then to determine which MDP to use and in which context, i.e. an MDP needs to be associated with its context of use. This is the goal of the Long Term Memory. 

\begin{figure}[htb!]
\begin{center}
    \includegraphics[width=0.8\linewidth]{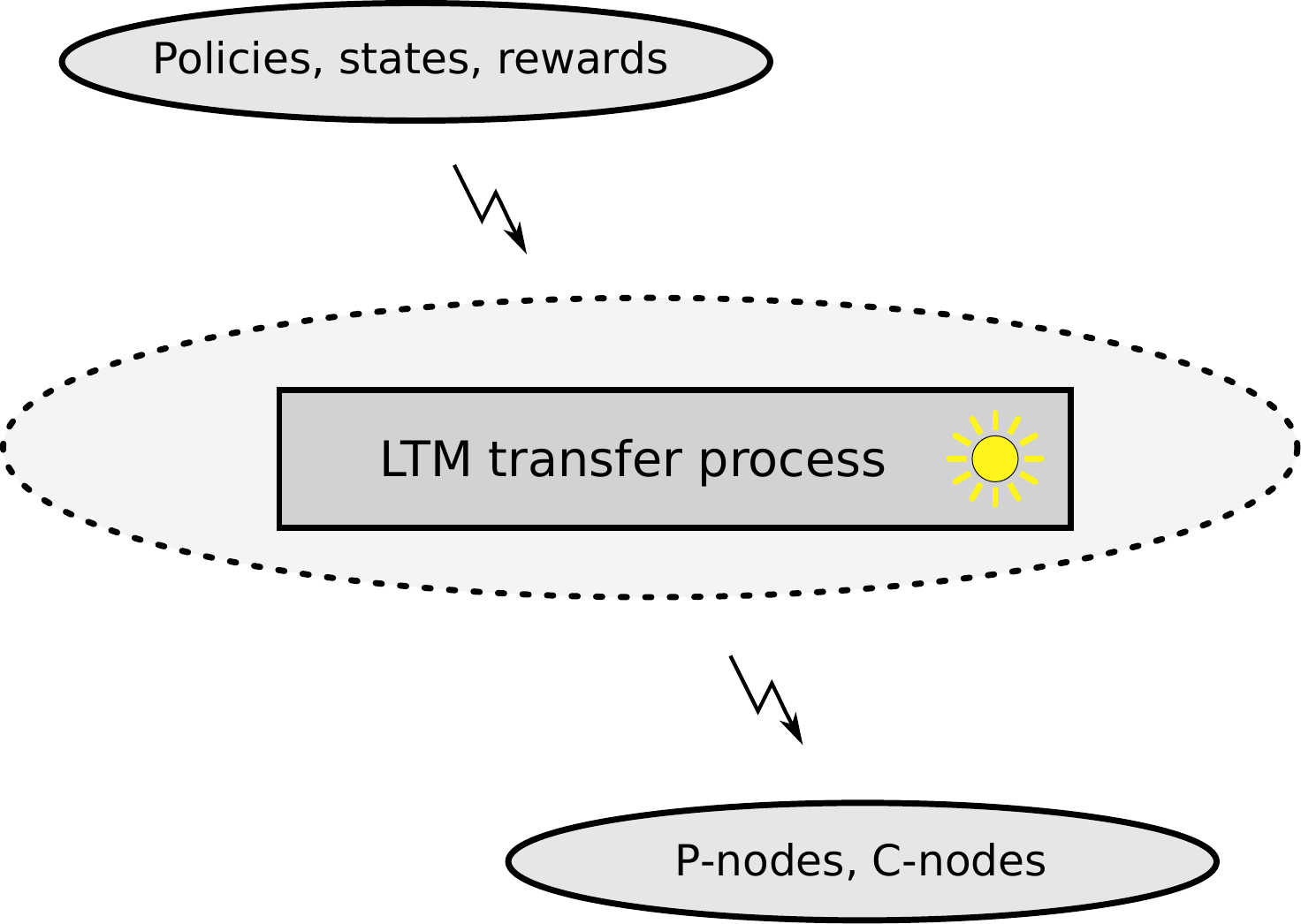}
\end{center}
\caption{\label{fig:STM_to_LTM}Transfer from short term memory to long term memory. Starting from policies, state spaces and rewards, the transfer process applies the policies and observes their outcome to deduce their context of use in the form of P-Nodes (perception nodes) and C-Nodes (context nodes).}
\end{figure}

\begin{figure*}[htb!]
	\begin{center}
		 \includegraphics[width=0.9\linewidth]{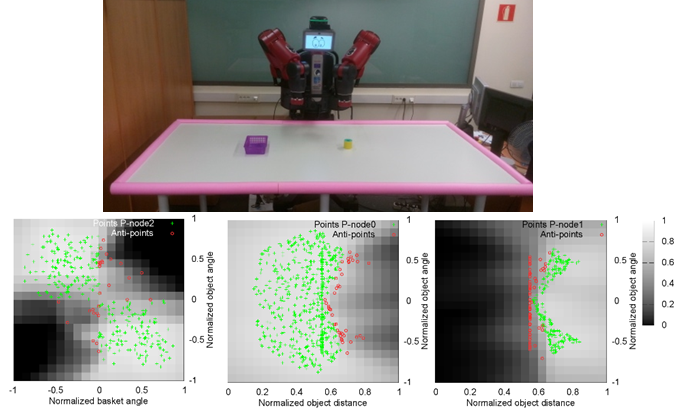}
	\end{center}
	\caption{\label{fig:wp3pnodes} Final activation maps for three P-nodes in an experiment where the robot tries to put randomly placed pucks into randomly placed baskets as shown in the setup on top.} 
\end{figure*} 

Long Term Memory is a fundamental part of any cognitive architecture that aims to store and reuse the acquired knowledge. Most traditional cognitive architectures (e.g. ACT-R, CLARION, SOAR) are based on symbolic representations and, thus, implement relatively straightforward LTM structures where reuse is based on searching for the appropriate labels. When working with subsymbolic structures, such as ANN based representations, labeling is not straightforward and other types of LTM approaches must be used. In this work, we have developed an experience-based associative LTM structure. The basic features of this approach are presented in \cite{duro2018perceptual}. The operation of this type of LTM is based on the relationships the system acquires about knowledge nuggets as it interacts with the world. The basic knowledge nuggets that are considered here are the different parts of the MDP, i.e. states, rewards and policies. The idea is that when instances of these knowledge nuggets co-occur and when something relevant was experienced by the system (such as a high reward), they are associated through a new type of knowledge nugget: a Context node (C-node). The definition and operation of these structures have been described in \cite{duro2017context}. C-nodes implement conjunctive representations of context and are activated when their associated nuggets are present (or at least most of them). In other words, a C-node implies that sets of the type $\{S_i, R_k, \pi_r\}$ that lead to relevant situations experienced by the system must be identified and stored. This way, when a sub-set of $\{S_i, R_k, \pi_r\}$ is active -- $S_i$ allows to predict what happens and a $R_k$ is activated --, the system can infer from its experience that applying policy \textit{$\pi_r$}, should lead to producing the same relevant event. These nodes provide a simple albeit powerful structure to store context related information that allows the system to selectively recall appropriate policies (or other knowledge nuggets) in the presence of known or similar situations.

Most of the knowledge nuggets stored in LTM have not usually been defined in the whole perception or state space. For instance, the accuracy of a reward function cannot be expected to be high far from observed areas of the state space. All knowledge nuggets in LTM, including C-nodes, are reliable only within a particular area of state space. Consequently, they should only be used or activated in this area. To address this issue, it is necessary to introduce the concept of perceptual classes, which are areas of the perceptual space for which knowledge nuggets are valid\footnote{Perceptual classes can also be used to represent other properties of knowledge nuggets but we focus on this one here.}.  Therefore, a perceptual class is a generalization of perceptions into a higher level, discrete, representation linked to a given response of the system. Perceptual classes are represented with a LTM component called perceptual nodes or P-nodes. A P-node is a functional component that is activated when a perceptual state belongs to a given perceptual class.

Different algorithms for the online and offline delimitation of P-nodes have been proposed, both using point-wise distance based representations, that is, heuristic episode clustering approaches \cite{duro2018perceptual}, and neural network based generalizations. Redescription procedures to go from a more hippocampal-like episode-based representation to a more cortex-like generalized representation in the form of ANNs were studied as reflected in \cite{becerra2018redescriptive}. This process can be carried out in a quasi-online manner with a reasonable quality level and it can also take place during an off-line dreaming-like process leading to much better results. For instance, \figurename~\ref{fig:wp3pnodes} displays a representation in the form of 2D activation maps of some P-nodes (representing perceptual classes) that were automatically obtained using a Baxter robot that was trying to learn to put objects placed anywhere into baskets, also placed anywhere. The bottom left graph represents one of the P-nodes and maps the angle at which an object is located with respect to the angle of the target basket to put the object. It can be interpreted as object in the wrong side (need to change hands). The bottom central graph maps distance and angle at which an object is located with respect to the robot, and its activation can be interpreted as “the object is reachable” as it provides the reachable area for the robot arm. Finally, the bottom right graph corresponds to non-reachable area or “unreachable object”.

Using this type of associative memory, after the system has acquired some experience interacting with an environment or a set of environments, whenever it is faced with a perceptual context, a set of P-nodes becomes active, thus pre-activating all of the knowledge nuggets that might be relevant in that situation in terms of perceptual cues. This provides an opportunistic way of pre-selecting previously learnt knowledge nuggets that might be relevant in order to execute them or to generate new knowledge nuggets for a new environment the robot might be facing.

\subsection{From one agent to another}

%

So far, we have limited our scope to individual robot learning. However, due to the variety of situations, exploration can be tedious. As stated in Section~\ref{sec:goalsandchallenges}, one multi-task challenge is to transfer acquired knowledge from one robot to another, so as to enable improvement over the experience of others. Indeed, using multiple robots can increase the efficiency with respect to both speed and quality of learning by sharing experience obtained by multiple exploration processes running in parallel. Here we consider robot-to-robot learning where multiple robots share learned skills while completing a task.

Robot-to-robot learning raises its own technical limitations (e.g.: communication bandwidth, network structure) and challenges (what and how much should be transferred, and to whom). Also, the problem of sharing information can be seen as a combination of both an exploration problem and a consensus problem. As described earlier, the exploration problem is addressed through individual learning, which is performed independently from the group. As a consequence, the consensus problem must comply with skills learned by different robots, where skills compete to be transferred to the whole group. Some robots may acquire and share better skills than other robots, and the question is open as to how to select the best skills while maintaining a certain level of diversity resulting in the discovery of even better skills.

We have explored two similar classes of algorithms for information sharing in multi-robot systems: embodied evolution~\cite{bredeche2018ee} and social learning~\cite{heinerman2015}. While the former emphasizes learning of collective behaviour (i.e. robots interacting with one another), the latter is explicitly concerned with sharing chunks of information that have been acquired by individual robots. However, we have shown that with both families, learning converges towards a homogeneous set of skills shared by the whole group of robots.

\begin{figure}[htb!]
\begin{center}
\includegraphics[width=\linewidth]{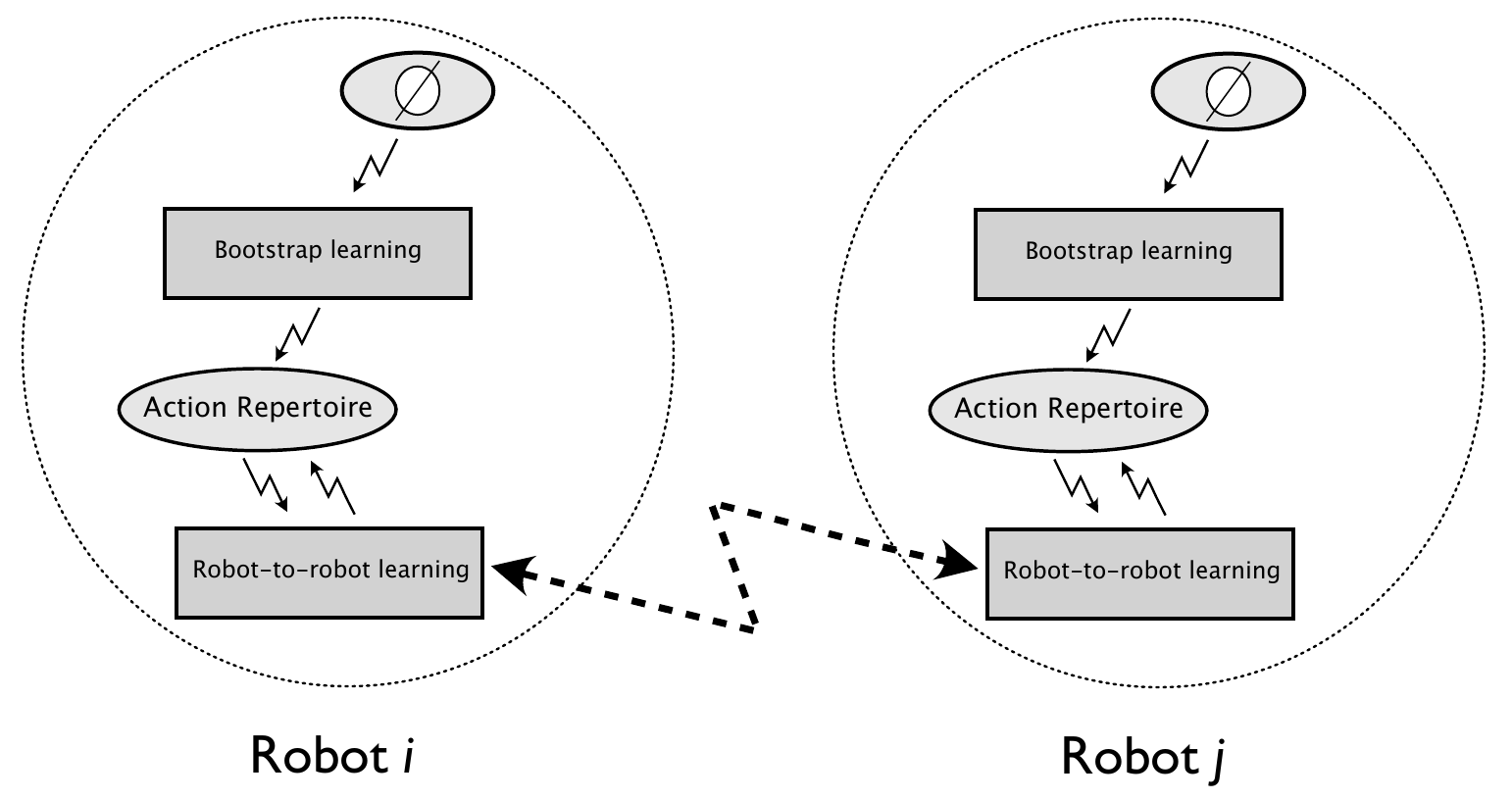}    
\end{center}
\caption{\label{fig:multirobots-transfer}Schematic view of robot-to-robot learning. First, each robot builds its own repertoire of actions by individual learning from experience ("bootstrap learning", see \figurename~\ref{fig:bootstrap} for a detailed view). Then elements from the repertoire of actions can be transferred from one robot to another, and vice versa. Both processes run in parallel.}
\end{figure}

The general architecture is illustrated in \figurename~\ref{fig:multirobots-transfer}. Each individual robot learns individually and transmits all or part of the description of its skill set to other robots. Information transfer depends on network connectivity: a given robot may broadcast to everyone, or to the subset of reachable robots. As with social learning in nature, selecting which set of skills is to be transferred and accepted depends on a selection process running on each robot. The more exclusive the selection, the faster the convergence, but at the cost of a faster loss of diversity in the collective. We have shown in~\cite{montanier2016spec,bredeche2017wk} that selection of incoming skills proportionally to their accounted performances provides an efficient way to maintain diversity of individual learning processes.

Even more importantly, we have also shown that, as expected, social learning provides increased learning speed, but also yields increased performance when compared to individual learning~\cite{heinerman2019}. This is due to the possibility of running several instances of individual learning algorithm with different meta-parameters values, as best values cannot be guessed before run-time. In other words, social learning can efficiently mitigate the negative effect of parameter tuning of the learning process by enabling multiple searches and selecting the best performing one at run-time. In addition, the gain in diversity can actually help to obtain even better results than those that could be obtained by the best robot learner alone.

\section{Related work}

The DREAM architecture aims at going beyond a single learning and decision process and thus moves towards cognitive architectures that are designed to coordinate such processes. In particular, it relies on a hierarchical architecture where higher representational levels are built on top of more elementary ones, and it targets the open-ended learning context, which is closely related to continual learning and developmental processes. In this section we investigate the relationship between our work and all these areas of research.

\subsection{Cognitive architectures}

Cognitive architectures have been studied for more than 40 years \cite{kotseruba201840}. One of the main goals of cognitive architecture research is to model the Human mind and understand it with a synthetic methodology. It has thus a wider scope than what is proposed here. These architectures are often classified according the kind of representations they can manipulate \cite{ye2018survey,kotseruba201840}. They are either:
\begin{itemize}
\item \textit{Symbolic:} cognitive architectures, relying on symbols and a dedicated instruction set, as GLAIR \cite{shapiro2010glair}, EPIC \cite{kieras2016summary} or ICARUS \cite{langley2006unified}. Knowledge nuggets in this case are generally represented as IF-THEN rules;
\item \textit{Emergent:} cognitive architectures, relying on connectionnist approaches, as BECCA \cite{rohrer2012becca}, MDB \cite{bellas2010multilevel} or SASE \cite{weng2004developmental}. The knowledge is distributed in neural networks;
\item \textit{Hybrid:} cognitive architectures, relying on both, as ACT-R \cite{anderson2004integrated} or SOAR \cite{laird2012soar}, that were initially symbolic, but that include non-symbolic representations, at least in their latest versions \cite{laird2012soar}.
\end{itemize}

Each architecture has it own processes to acquire new knowledge, but the knowledge representation is a core feature that is given beforehand. For practical reasons, it is generally homogeneous \cite{lieto2017representational} and considered as a design choice that the system cannot act on. As such, representational redescription is not addressed in these works. Sometimes, it is even difficult to figure out what type of representation is used, resulting in different classifications of cognitive architectures between different review papers \cite{kotseruba201840}. 

The DREAM architecture is not a full cognitive architecture. It is a consistent set of modules aimed at providing robotic cognitive architectures with a new ability: the ability to autonomously build new knowledge representations that are adapted to robot's tasks and features. It can be seen as an instance of the \textit{design by use case} approach proposed by \cite{vernon2017two}.

\subsection{End-to-end and hierarchical approaches}

The ability to deal with multiple representations appears in the machine learning literature under the point of view of hierarchical approaches able to represent available knowledge at multiple levels. It relies on the notion of \textit{options}, that extends the mathematical framework of Markov Decision Processes by adding different levels of temporal abstractions. The Markov Decision Process becomes a semi-Markov Decision Process, with actions that may have different temporal extensions \cite{sutton1999between}. In this Hierarchical Reinforcement Learning (HRL) framework, options are defined as a triplet $\langle \mathcal{I}, \pi, \beta \rangle$, where $\mathcal{I}$ is the initiation set, i.e. the set of states from which the option can be activated, $\pi$ is the policy selecting the action to perform and $\beta$ is the termination condition. With this approach, the combination of different levels of representations is possible. It can accelerate learning \cite{mann2014scaling} or make learning more robust \cite{mankowitz2018learning}.

In the standard option framework, all options are built on top of an initially provided MDP. As a result, the robot controlled by such hierarchical approaches is limited by this initial MDP. Given the difficulty to define an MDP in a robotics context \cite{kober2013reinforcement}, this is a strong limitation to implement versatile robot learning capabilities.

Deep reinforcement learning holds the promise of circumventing this difficulty by making it possible to learn from raw inputs to raw outputs \cite{arulkumaran2017brief}. With such end-to-end approaches, it becomes possible to build policies taking images as inputs and generating raw motor commands, with applications in grasping \cite{levine2016end} or self-driving cars \cite{bojarski2016end}, for instance. However, these approaches are still limited. For the grasping experiment, full observability is required \cite{levine2016end}. The considered tasks also require to move the robot end effector to positions that are known and this knowledge is taken into account in the cost functions used. Current approaches thus need to be completed with modules in charge of preparing the data they need. This is one of the goal of the proposed approach. Besides, these approaches often require very large data sets which makes them generally impractical. The self-driving car application, for instance, relies on 72 hours of driving data \cite{bojarski2016end}.

This issue is addressed in a variety of ways, such as learning a lower dimensionality representation \cite{bohmer2015autonomous,Lesort18}, or using learned models of the environment \cite{chatzilygeroudis2019survey}. But to keep with the hierarchical learning perspective, standard HRL methods have been extended to the deep learning context with various frameworks \cite{bacon2017option,vezhnevets2017feudal,levy2017hierarchical,levy2018hierarchical,nachum2018data}. However, it is still hard to find application of these frameworks to a real robot, except when crossing the reality gap is straightforward \cite{yang2018hierarchical,chen2019attention}.

Abstract policies can also be extracted from a set of demonstrations \cite{rahmatizadeh2018vision} or at least "informative" policies that may result from a former training session \cite{fox2017multi,krishnan2017ddco}. These works assume that such policies are available or that informative policies can be learned. In an open-ended learning scenario, the acquisition of a relevant data set should be included in the learning process itself to make the system more autonomous. Our approach aims at building such policies in realistic setups with sparse reward and is thus complementary with them. Other approaches avoid this issue by adding intrinsic motivations \cite{vigorito2010intrinsically,kulkarni2016hierarchical,florensa2017stochastic}. Again, only few of them have been applied to a real robot \cite{forestier2017intrinsically}, and they have not been connected to representational redescription concerns yet, even if some preliminary work comes into this direction \cite{Pere2018,laversanne2018curiosity}.

\subsection{Continual learning and development}

Hierarchical approaches are focused on structuring policy representations to better exploit what has been learned on a new task. Tasks are, in general, implicitly related and in a limited number. When learning over long time periods and many different tasks, new issues arise: training data cannot be completely saved and the number of learned policies increases. The consequence is a risk of {\em catastrophic forgetting} or the difficulty to identify relevant stored policies. Approaches dealing with these issues have been given different names: lifelong learning \cite{thrun1998lifelong, parisi2019continual}, never ending learning \cite{mitchell2018never} or continual learning \cite{lesort2020continual}.  With the transfer between short term and long term memory, our approach includes an instance of dual-memory learning systems \cite{parisi2019continual}. However, while the focus in these works is in building learning processes that can deal with the continuous flow of data and the different tasks with a single learning process \cite{lesort2020continual}, our approach decomposes learning into different processes in charge of bootstrapping the representational redescription process, acquiring skills and consolidating them to make them more robust and transferrable between domains, tasks and robots. 

The decomposition of the learning process into different phases is a feature of developmental robotics \cite{lungarella2003developmental,weng2004developmental,cangelosi2015developmental}, that draws inpiration from human and animal development. In these approaches, the robot is not ready to solve a task when it is first turned on. It needs to acquire first information about itself and its environment. During this phase, no task is considered. The robot is just exploring, with dedicated intrinsic motivations \cite{oudeyer2007intrinsic, oudeyer2009intrinsic, baldassarre2013intrinsically, oudeyer2018computational}, to identify what is possible and generate data to learn models of the world (including itself) and sensorimotor skills. This is a fundamental difference with continual learning approaches where the robot does nothing else than solving a task from the very first moment it is turned on. The approach we have proposed is focused on the acquisition of adapted representations, it is thus an \textit{early developmental AI} system according to the classification of \cite{guerin2011learning}.

Pioneering works on this topic drew inspiration from Piaget's developmental Psychology work and applied it to simplified environments with predefined representations \cite{drescher1991made,chaput2004constructivist}. Later works focus on the question of building appropriate discrete representations from low-level sensorimotor values \cite{stober2008pixels, mugan2011autonomous,zimmer2017bootstrapping}, some even going towards abstract symbolic representations \cite{konidaris2014constructing,konidaris2016constructing}. While these works do consider simulated robots, the authors of \cite{ugur2015bottom} propose a method to build abstract representations that has been tested on real robots. It starts from a fixed set of known actions and thus does not address the skill discovery challenge. 

To summarize, many different challenges have to be faced when trying to apply machine learning methods to real robots (see Section  \ref{sec:open-ended_learning}). There are some approaches combining deep HRL and lifelong learning mechanisms \cite{tessler2017deep} but, to the best of our knowledge, the approach introduced here is the first one that considers most of them, lists and tests relevant approaches and describes how to connect and articulate them while applying it to real robots. 

\section{Impact on Neuroscience}
\label{sec:disc_neuro}
So far, we have shown the importance of representational redescription from a robotics perspective, arguing that it is a critical process for the versatility of robots in realistic open-ended learning scenarios. But representational redescription is above all a key process in the cognitive capabilities of living creatures. In this section, we investigate what Neuroscience can learn from our work.

\subsection{Neuroscience and state representation redescription}

The hippocampus is well known, even in the machine learning and robotics communities, for hosting neurons whose receptive fields appear to represent locations, the so-called \emph{place cells} \cite{okeefe1971}. Many models of the hippocampus have been implemented on real robots \cite[for example]{arleo2000spatial,krichmar2005spatial,strosslin2005robust,giovannangeli2006robustness,barrera2008biologically} or have inspired robot navigation algorithms \cite[for example]{milford2010,caluwaerts12}, either to test the efficiency of neural theories in realistic settings, or as bioinspired tools for robotic navigation. 
The resulting compact and efficient representation of spatial states has then often been used to learn navigation behaviors by reinforcement \cite[for example]{guazzelli1998affordances,arleo2000spatial,foster2000model,chavarriaga2005computational,girard2005integration,caluwaerts12}. In particular, this has been done by connecting models of the hippocampus to a model of the basal ganglia (a group of subcortical nuclei known to be involved in action selection) in order to generate goal-directed behaviors. This fits quite well with the traditional machine learning approach where one designs a state representation (here, the spatial position) adapted to the task one wants to solve (here navigation), and then use a reinforcement learning algorithm to learn the optimal policy.

However, recent experimental Neuroscience results shed a new light on what the hippocampus might in fact be doing: rather than only representing places, it may in fact encode the general representation that is the most appropriate to handle the task at hand, even when the task is not spatial (e.g., categorization of social agents \cite{tavares2015,park2019}). This representation would correspond to 2D localization when reward delivery is driven by the ability to reach a given location in space, but it could also be the integration of time and/or distance \cite{kraus2013hippocampal}, or the position in a sequence \cite{cabral2014oscillatory}, if these variables are the essential ones to earn reward. 
This suggests that some reinforcement signals are used in the hippocampus so as to sort out which of the few dimensions that can be extracted from the sensory data are relevant for the task at hand: encoding durations, distances or sequence order, only if using them helps getting rewards/not using them leads to poorer performance.

These observations fit quite well with the approach advocated here for learning in robotics: the hippocampus may indeed be a central player, when it comes to performing \emph{representational redescription}. 


Actually, the algorithms proposed here could be used to derive robotics-informed hypotheses about how the brain might perform representational redescription. For instance, in Section~\ref{sec:state_repr}, we showed that an efficient State Representation Learning (SRL) algorithm must combine several different objectives into the loss function: reconstructing observations (e.g., learning to encode both target and robot positions), learning forward and inverse models, and predicting reward. In particular, we found that without these complementary learning objectives, predicting reward leads only to a classifier detecting when the robot is at the goal, thus not providing any particular structure or disentanglement to the rest of the state space.

In contrast, recent Neuroscience work attempting to model adaptive state creation during animal reward learning generally focused on too restricted representations \cite{redish2007,gershman2010}. We argue that some knowledge of the SRL work presented in this paper may help them better understand how animals perform representation learning when facing tasks represented with a larger number of dimensions.

\subsection{Neuroscience and task-sets coordination}

The present work may also have implications for another field of neuroscience research, namely the study of task-set learning and coordination in the prefrontal cortex \cite{miller2001,collins2012}. Learning different task-sets means learning different sets of action values (e.g., different Q-tables) in parallel and adaptively shifting to the set which seems the most appropriate for the task at hand. This process has also been called ``
episodic control''. 
In these tasks, Human subjects need to perform various cognitive operations: autonomously detect that the task has changed though these changes are not signalled, store in memory the set of action values associated to the previous task, search in memory whether there already exists a previously learned set of action values that corresponds to the new task, or instead learn the new action values, and so on after each abrupt task change. 


Importantly, computational models of task-set learning typically use the same state/action descriptions. The difference between task-sets thus relies in the mapping between states and actions (i.e., a different Q-table per task-set). The focus of these models is on how the prefrontal cortex detects task context changes to decide which memorized (previously learned) task-set is now relevant, or whether the situation is novel and a new task-set should be created and learned. Now, the work presented in the present paper proposes to go beyond this, by adding the possibility of ``representational redescription''. In other words, different tasks may not only require different action values for the same state/action representation, but also sometimes different state or action representations.

One interesting question is then: are prefrontal cortex mechanisms for state/action representational redescription completely different from those used to change/coordinate task-sets? 
Interestingly, neural representations of different cognitive tasks can be learned sequentially in a continual learning setting; and it seems that the compositionality they give rise to could be used as part of the redescription or recombination strategies \cite{yang2019}. These novel questions open the road for further research at the crossroads between machine learning, autonomous robotics and computational Neuroscience.

\subsection{Neuroscience and action representation redescription}

A third important link which can be drawn with neuroscience research is about action representation redescription. More precisely, to our knowledge, most cognitive Neuroscience researches does not address the question of how novel action representations emerge in the nervous system. Instead, most assume that action representations are already in place, and focus on the question of how an agent can learn to select appropriate actions at a given moment. This action selection problem has received a lot of attention since several decades, and appears to involve the basal ganglia, a group of subcortical nuclei involved in the temporal organization of motor decisions \cite{redgrave1999,khamassi2005,khamassi2012}.

Nevertheless, interestingly, the basal ganglia has been found to also contribute to an action chunking mechanism resulting in the encoding of novel macro-actions constituted of a sequence of existing unitary actions \cite{graybiel1998,barnes2005,jin2014}. The rationale is that actions that are often repeated one after another within the same sequence (\textit{e.g.,} grasping a bottle, lifting the bottle, bending the bottle, pouring water into a glass, putting the bottle back on the table) can become a chunked routine that can then be triggered as a unitary habitual behavior when faced with the same context and goal \cite{graybiel1998,dezfouli2012,miller2019,morris2019}. Mechanistically, it is thought that neural activity related to each individual action, when sequentially activated within short periods of time, can be associated through Hebbian learning within the motor cortex \cite{ashby2010,frank2006}, so that after habit acquisition basal ganglia neurons become only active at the beginning of the sequence (putatively to initiate it) and at its end (putatively to indicate that it's finished) \cite{thorn2010,jin2010}.

Importantly, the action representation redescription experimental results presented here go further simple action chunking mechanisms for habit learning, and thus have the potential to raise novel insightful ideas for Neuroscience.

\subsection{Neuroscience \& sleep-related learning processes}

A last piece of results which may have impacts on Neuroscience, is about sleep-related learning processes.
As we previously mentioned, within the mammal brain, the hippocampus contains neurons which encode specific locations of the environment, the so-called ``place cells'' \cite{okeefe1971}. Strikingly, hippocampal place cells are reactivated during sleep while an animal is immobile and may be ``dreaming'' about the task that it previously performed \cite{wilson1994}. Such a replay not only occurs during sleep, but also during periods of quite wakefulness where the animal seems to be thinking about what it just did during a task \cite{foster2006,gupta2010}. Neuroscientists wonder what might be the role of such a replay phenomenon. One recent hypothesis is that such a replay might be useful to bootstrap reinforcement learning by using an internal model of the task structure to update state-action value functions offline \cite{mattar2018,khamassi2020}, similarly to the Dyna architecture \cite{sutton1991}.

Nevertheless, applying this type of neuro-inspired replay models to continuous state space navigation, which constitutes a step towards implementing them in real robots, revealed that offline replay of these models is not only useful to bootstrap the state-action value function, but also to learn a stable model of the world \cite{aubin2018,caze2018hippocampal}. More precisely, when solving a given navigation problem, the states are often encountered one after another, almost always in the same order. Online learning with a neural network is disrupted by such temporal correlations of the samples \cite{mnih2015}. Replaying past experiences in a random order was thus necessary to break these repeating temporal correlations, so as to learn a stable and coherent model of the world.



This last example gives us the opportunity to highlight a more general impact that robotics research may have on Neuroscience: because robots have to interact with the real world, testing learning algorithms in robots often leads to different results than perfectly controlled simulation models \cite{khamassi2011}. Some models that work perfectly well in simulations can lead to disappointing results when tested on a real robot. Alternatively, some models that appear suboptimal compared to other models in simulation may turn out more robust when facing noisy, multidimensional, sometimes unpredictable situations to which a robot is often confronted. Finally, some models that give the best performance in simulation may turn out computationnally too costly to work in real time on a robot. We thus hope that the open-ended learning robotics results presented in this paper, and thoroughly discussed above, can help convince neuroscientists of the interest of paying attention to robotics research, in addition to dematerialized work done in artificial intelligence, to better understand the properties of learning processes that can work in the real-world in a variety of experimental contexts.


\section{Conclusion}

The open-ended learning capability corresponds to the ability to solve tasks without having been prepared for them. A strong requirement to reach this capability is to give the robot the ability to build, on its own, the representations adapted to these tasks, may it be state or action spaces. We have discussed the challenges it raises and proposed the DREAM architecture, an asymptotically end-to-end, modular, and developmental framework to address them, emphasizing "awake" processes, that require the robot to interact with its environment and have thus a high cost, and "dreaming" processes, that do not. A partial implementation of this framework has been presented together with the results it has generated. Future work will complete this implementation to reach a complete open-ended learning ability.

\section{Acknowledgments}

This work has been supported by the FET project DREAM\footnote{\url{http://dream.isir.upmc.fr/}}, that has received funding from the European Union's Horizon 2020 research and innovation programme under grant agreement No 640891.

\bibliographystyle{unsrt}
\bibliography{biblio.bib}

\end{document}